\documentclass[journal]{IEEEtran}
\usepackage{ifpdf}

\usepackage{cite}
\usepackage{hyperref}

\ifCLASSINFOpdf
   \usepackage[pdftex]{graphicx}
\else
\fi
\usepackage{amsmath}
\usepackage{algorithmic}

\usepackage{array}
\usepackage{url}

\usepackage{booktabs} 
\usepackage{amsfonts,amssymb}
\usepackage{subfigure}
\usepackage{color,xcolor}
\usepackage{times}
\usepackage{epsfig}
\usepackage{amsmath}
\usepackage{textcomp}
\usepackage{gensymb}
\usepackage{wasysym}
\usepackage{multirow}
\usepackage{cite}
\usepackage{graphicx} 
\usepackage{colortbl}
\usepackage{algorithm}
\usepackage{algorithmic}

\usepackage{mathrsfs}
\usepackage{url}
\usepackage{lettrine}

\definecolor{mycyan}{cmyk}{.3,0,0,0}

\def\eg{\emph{e.g.}}
\def\ie{\emph{i.e.}}
\def\etc{\emph{etc}}

\begin{document}
%
\title{Hierarchical Perceptual Noise Injection for Social Media Fingerprint Privacy Protection}
%
%
%
\author{Simin Li, Huangxinxin Xu, Jiakai Wang, Aishan Liu\textsuperscript{\dag}, Fazhi He\textsuperscript{\dag}, Xianglong Liu, and Dacheng Tao, \emph{Fellow, IEEE}


\thanks{S. Li, J. Wang, A. Liu and X, Liu are with the State Key Lab of Software Development Environment, Beihang University, Beijing 100191, China. X. Liu is also with Beijing Advanced Innovation Center for Big Data-Based Precision Medicine. H. Xu and F. He are with the School of Computer Science, Wuhan University. D. Tao (Fellow, IEEE) is the Inaugural Director of the JD Explore Academy and a Senior Vice President of JD.com. He is also an Advisor and a Chief Scientist of the Digital Sciences Initiative, The University of Sydney. (\dag\ Corresponding author: Aishan Liu, liuaishan@buaa.edu.cn and Fazhi He, fzhe@whu.edu.cn)}
}



%
%

\markboth{IEEE Transactions on Image Processing,~Vol.~, No.~, July~2022}%
{Shell \MakeLowercase{\textit{Li et al.}}: Hierarchical Perceptual Noise Injection for Social Media Fingerprint Privacy Protection}
%



\maketitle

\begin{abstract}
Billions of people are sharing their daily life images on social media every day. However, their biometric information (e.g., fingerprint) could be easily stolen from these images. The threat of fingerprint leakage from social media raises a strong desire for anonymizing shared images while maintaining image qualities, since fingerprints act as a lifelong individual biometric password. To guard the fingerprint leakage, adversarial attack emerges as a solution by adding imperceptible perturbations on images. However, existing works are either weak in black-box transferability or appear unnatural. Motivated by visual perception hierarchy (i.e., high-level perception exploits model-shared semantics that transfer well across models while low-level perception extracts primitive stimulus and will cause high visual sensitivities given suspicious stimulus), we propose FingerSafe, a hierarchical perceptual protective noise injection framework to address the mentioned problems. For black-box transferability, we inject protective noises on fingerprint orientation field to perturb the model-shared high-level semantics (\ie, fingerprint ridges). Considering visual naturalness, we suppress the low-level local contrast stimulus by regularizing the response of Lateral Geniculate Nucleus. Our FingerSafe is the first to provide feasible fingerprint protection in both digital (up to 94.12\%) and realistic scenarios (Twitter and Facebook, up to 68.75\%). ~\footnote{Our code can be found at \url{https://github.com/nlsde-safety-team/FingerSafe}.}

\end{abstract}

\begin{IEEEkeywords}
Fingerprint, Adversarial attack, Privacy protection.
\end{IEEEkeywords}

%
\IEEEpeerreviewmaketitle

\section{Introduction}
\begin{figure}[t]
\centering
\includegraphics[scale=0.35]{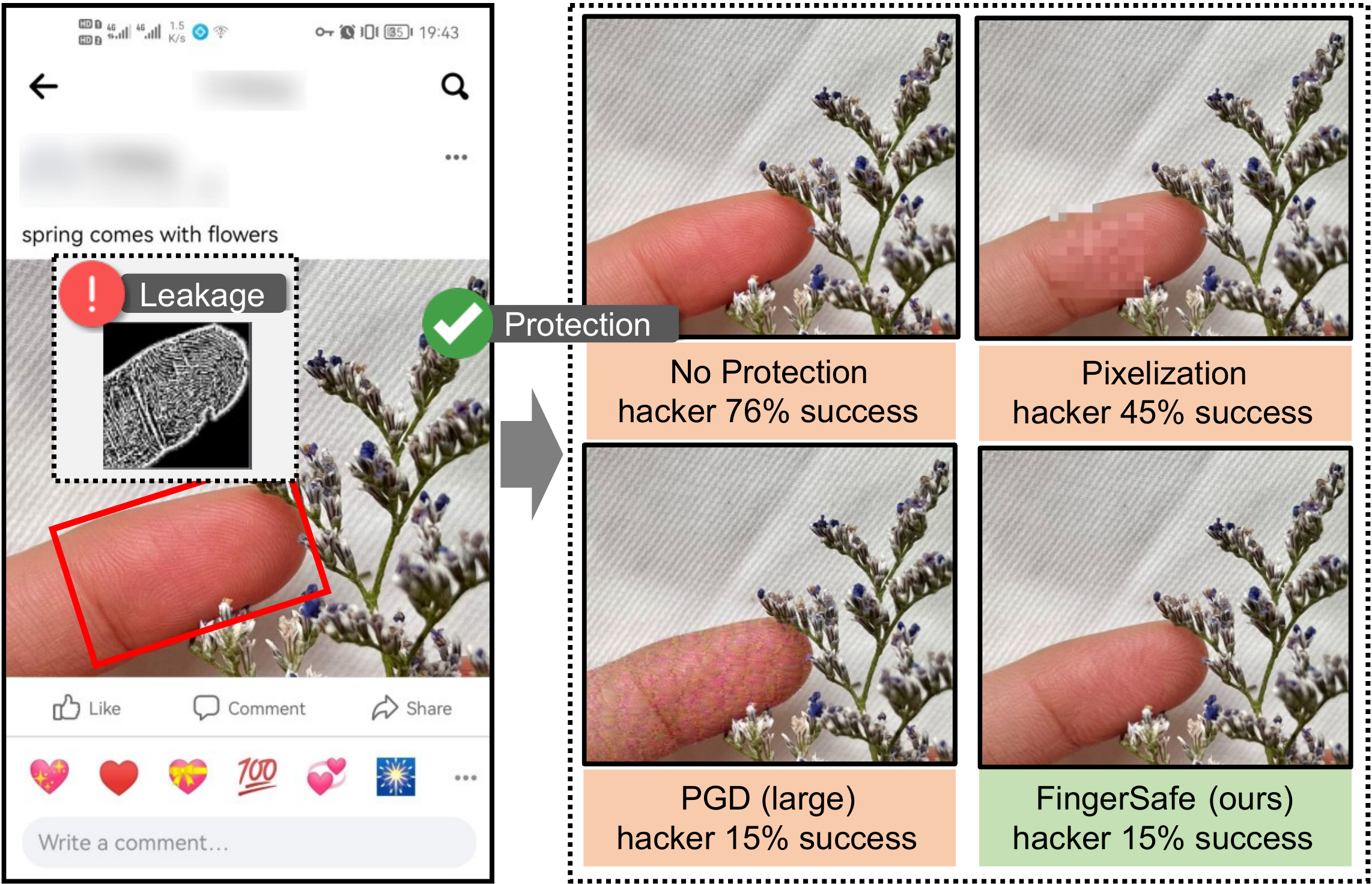}
\caption{Posts on Facebook unconsciously leak fingerprint of owners. The protection of pixelization is limited, large PGD is effective but unnatural. In contrast, FingerSafe is effective and natural. Images used with consent from owners.}
\label{intro}
\end{figure}

\IEEEPARstart{P}osting photos on social media is a popular way to share our daily life. However, with thousands of publicly shared images on Instagram containing accessible fingerprint details \cite{malhotra2020cvprw}, personal biometric information (\eg, fingerprint) can be easily stolen from the photo shared on social media, which may cause severe security problems to fingerprint authentication systems (\eg, access control system, FingerID) as shown in Fig.\ref{intro}. Extensive evidences have revealed the feasibility of the above challenges, for example, hackers easily accessed the fingerprint of the president of EU commission via an online photo in 2014 \cite{hern2014hacker}; recently, fingerprints are easily extracted from images shared on social media \cite{jo2017leakfinger,CNN2018leakdrug,CNN2021leakcheese}, even from images photoed 3 meters away \cite{harding2017ninefeet}. The leakage of fingerprint is \emph{irreversible}---since you cannot change your fingerprint, once the fingerprint is leaked, all systems that rely on fingerprints are at risk for the rest of your life \cite{wei2015lifetime}. Based on these leaked fingerprint images, hackers can sneak in access control systems of governments, banks and the police \cite{zak2019threatofleakage}, or apple pay \cite{silicone} with more than 80\% success rate with an inkjet-printed paper \cite{cao2016hackusingpaper} or 3D printed mold \cite{rascagneres20203dmold}. \emph{The fingerprint security is at severe risk now.}

Wiping out all fingerprint details in an image seems to be a simple solution to protect fingerprint privacy (\eg, masking or pixelization). However, images protected by such methods are highly unnatural, often inappropriate to share on social media (as shown in Fig.\ref{intro}). Moreover, hiding fingerprint details by ad-hoc de-identification methods, such as blurring, is of weak protection performance \cite{gross2006model}. Recently, adversarial attacks \cite{szegedy2013intriguing,goodfellow2014explaining} have been used to protect personal privacy by adding small perturbations that are imperceptible to humans but misleading to DNNs \cite{ICLR2021unlearnable,shan2020fawkes,yang2020JunZhu,malhotra2020cvprw,cherepanova2021lowkey}.

However, several challenges still remain towards real-world fingerprint privacy protection: (1) \emph{Failure in black-box transferability}. Current studies primarily focus on white-box privacy protection, which assumes that all information of target recognition model used by attacker is known to protector. In real world, attackers might use all kinds of recognition models for the stolen fingerprint, while succeed on any model means failure in privacy protection. Thus, the information of attacker model is not known \emph{a priori} to the protector, resulting in a black-box model. As a result, the protection method must be effective towards different purposes (\eg, identification, verification) with different model architectures (\eg, DNN, traditional non-learning methods) using different preprocessing algorithms of attackers. (2) \emph{Undesirable visual naturalness}. Existing effective protection strategies do not yield natural appearance. The added perturbations are suspicious to human vision, making users reluctant to share such strange and unnatural images on social media.



Biological studies \cite{groen2017lowtohigh} revealed that visual perception process is often formulated via a low-to-high hierarchy. Perception stages at different levels focus on different abstract features and have different goals. High-level perception focuses on predictions using abstract semantics that are shared between models \cite{groen2017lowtohigh,wang2021dualattention}, while low-level perception extracts primitive features and alerts unnatural appearance (high visual sensitivities) when suspicious stimulus presents \cite{berman2014plosone,groen2017lowtohigh}. Motivated by these observations, this paper proposes a hierarchical noise injection framework named FingerSafe to address the above problems. As for \textbf{black-box transferability}, we attack high-level ridge features (the unique patterns on fingertip \cite{hong1998tpami}) by injecting protective noise on fingerprint orientation field (the pixelwise orientation of fingerprint correlated to ridge feature). Since most models in fingerprint recognition are built upon ridge features, FingerSafe achieves desirable transferability by perturbing such model-agnostic features. As for \textbf{visual naturalness}, local contrast (a low-level stimulus) triggers largest human response in unnaturalness \cite{berman2014plosone}. We therefore suppress the low-level local contrast using the response of ganglion cells of the retina and Lateral Geniculate Nucleus (LGN) in human visual system \cite{tadmor2000localcontrast} for visual naturalness. The \textbf{contributions} of this paper are summarized as follows:

\begin{itemize}

    \item To the best of our knowledge, FingerSafe is the first work to achieve strong black-box protection capability (\eg, unknown architectures, preprocessors, \etc) in fingerprint privacy protection.
    
    \item We propose orientation distortion module (high-level semantics) and local contrast suppression module (low-level stimulus), such that FingerSafe simultaneously achieves better result in transferability and naturalness.
    
    \item Extensive experiments demonstrate that FingerSafe outperforms baselines by large margins in digital and real-world scenarios (Twitter and Facebook).
\end{itemize}

\section{Related Works}
In this section, we briefly review related works in fingerprint recognition, adversarial attacks and privacy protection.

\subsection{Fingerprint Recognition}
Fingerprint recognition system primarily consists of four basic steps: capturing biometric fingerprint data, preprocessing, feature extraction and matching. Preprocessing algorithms are designed to keep ridge features of fingerprint and filter out noisy and spurious features \cite{jain1997tpamiminutiae,sankaran2015ijcb,lin2017ijcb}. Afterwards, discriminative fingerprint features were extracted from preprocessed images. Traditional methods are usually minutiae-based (\eg, ridge ending, bifurcation and short ridge) \cite{afsar2004fingerprint,cappelli2010minutia} or local descriptor-based \cite{zheng2015suspecting,malhotra2020scatnet,bruna2013scatnet}. Recently, neural network has been introduced as a feature extractor \cite{tang2017fingernet,darlow2017fingerprint,lin2018contactless}. Matching characterizes the purpose of recognition: for fingerprint verification, the goal is to determine if the two fingerprints are from the same finger \cite{jain1997tpamiminutiae,upmanyu2010blind}; for fingerprint identification, the goal is to search for a query fingerprint in the database\cite{sun2015deep,ali2016overview}. In this paper, we consider privacy protection in both fingerprint verification and identification tasks.

\subsection{Adversarial Attacks}
Adversarial examples are elaborately designed perturbations which are imperceptible to humans but could fool DNNs to wrong predictions \cite{goodfellow2014explaining,szegedy2013intriguing,liu2019psgan, wang2021universal}. Such vulnerability of machine learning models has raised great concerns in the community and many works have been proposed to improve the attack performance \cite{moosavi2016deepfool,carlini2017towards,papernot2016limitations} and search for possible defenses \cite{carlini2016defensive,liu2021ANP,samangouei2018defense,meng2017magnet, zhang2020interpreting, yu2021progressive}. In general, adversarial attack methods can be categorized into white-box \cite{madry2017pgd,shan2020fawkes,malhotra2020cvprw} (attackers have direct access to the structure, parameters of the model) and black-box attacks \cite{cheng2018query,yang2020JunZhu,wang2021dualattention} (attackers have limited knowledge on the model, \eg, unknown architecture and operations). In this paper, we primarily focus on black-box attacks, \ie, using adversarial attacks to protect fingerprint privacy in the black-box setting.

\subsection{Privacy Protection}
Privacy protection aims at making hackers impossible to exploit personal data from acquired content. A simple solution for privacy protection is to obfuscate the image, \eg, blurring, pixelation, darkening \cite{wilber2016can}. However, these methods are either ineffective \cite{mcpherson2016defeating,oh2016faceless} or visually unsatisfying \cite{malhotra2020cvprw}, which cannot be applied to protect fingerprint privacy on social media. Recently, several works have been devoted to protect biometric privacy using human-imperceptible noises. Unlearnable examples \cite{ICLR2021unlearnable} were proposed to protect privacy by fooling models to learn meaningless features during training. To protect facial images, Fawkes \cite{shan2020fawkes} was devoted to generate noises that create maximal changes in image representation. Similarly, LowKey \cite{cherepanova2021lowkey} used an ensemble strategy and added a loss term on Fawkes that additionally perturb image representation under Gaussian smoothing. TIP-IM \cite{yang2020JunZhu} protected facial images by generating adversarial identity masks comprised of many target identity. For fingerprint protection, \cite{malhotra2020cvprw} is the only research in white-box setting. However, their white-box protection performance is limited and hackers can sidestep such white-box protection by simply using another fingerprint recognition method. 

In contrast to previous studies, our FingerSafe achieves strong black-box transferability and naturalness, which could protect fingerprint privacy in complex real-world scenarios (\eg, hackers have different purposes, use different learning algorithms, model architectures, \etc).

\section{Approach}
\subsection{Problem Definition}
Given an image $\mathbf x$ containing fingerprint, a DNN model $\mathbb F$ is designed for fingerprint recognition on preprocessed fingerprint $\mathbf P\left (\mathbf x \right )$ such that $\mathbb F \left (\mathbf P \left (\mathbf x \right) \right ) = y$, where $\mathbf P$ is the image preprocessing algorithm, and $y$ is ground truth label of $\mathbf x$ (\ie, person ID).

In real-world, \emph{hackers} could leverage the leaked image $\mathbf x$ to sneak in the third-party FingerID system $\mathbb F_{\theta}$, where we call it \emph{testing stage attack}. Correspondingly, in \emph{testing stage protection}, we aim to protect the abuse of fingerprint privacy by the hackers. Technically, we generate adversarial examples $\mathbf{x}_{adv}$ that is visually similar to clean examples $\mathbf{x}$, but hackers cannot sneak in $\mathbb F_{\theta}$:
\begin{equation}
\mathbb F_{\theta} \left (\mathbf P \left (\mathbf x_{adv} \right) \right ) \neq y  \quad s.t. \quad \|\mathbf x-\mathbf x_{adv}\| \leq \epsilon,
\end{equation}
where $\| \cdot\|$ is a distance metric (\eg, $\ell_1, \ell_2, \ell_\infty$) to constrain the difference between $\mathbf{x}$ and $\mathbf{x}_{adv}$. In this paper, we call $\mathbf{x}_{adv}$ protected images, or adversarial examples.

In practice, protectors are not aware of the model used by the third-party authority (\ie, $\mathbb F_{\theta}$ is a black-box model), and different authorities might use different models as well. We thus generate the protected image $\mathbf x_{adv}$ from a surrogate model $\mathbb F_{\vartheta}$ and \emph{transfer} it to unknown target model $\mathbb F_{\theta}$.


In this paper, we focus on testing stage protection. However, we find FingerSafe is also effective in protecting \emph{training stage attacks}, where the goal of hackers is to collect $\mathbf x$ and use it to train a highly accurate model $\mathbb F_{\Theta}$ for commercial use. Note that $\mathbb F_{\Theta}$ is the model trained by hackers and $\mathbb F_{\theta}$ is used by an authorized third party.

\begin{figure*}[t]
\centering
\includegraphics[scale=0.6]{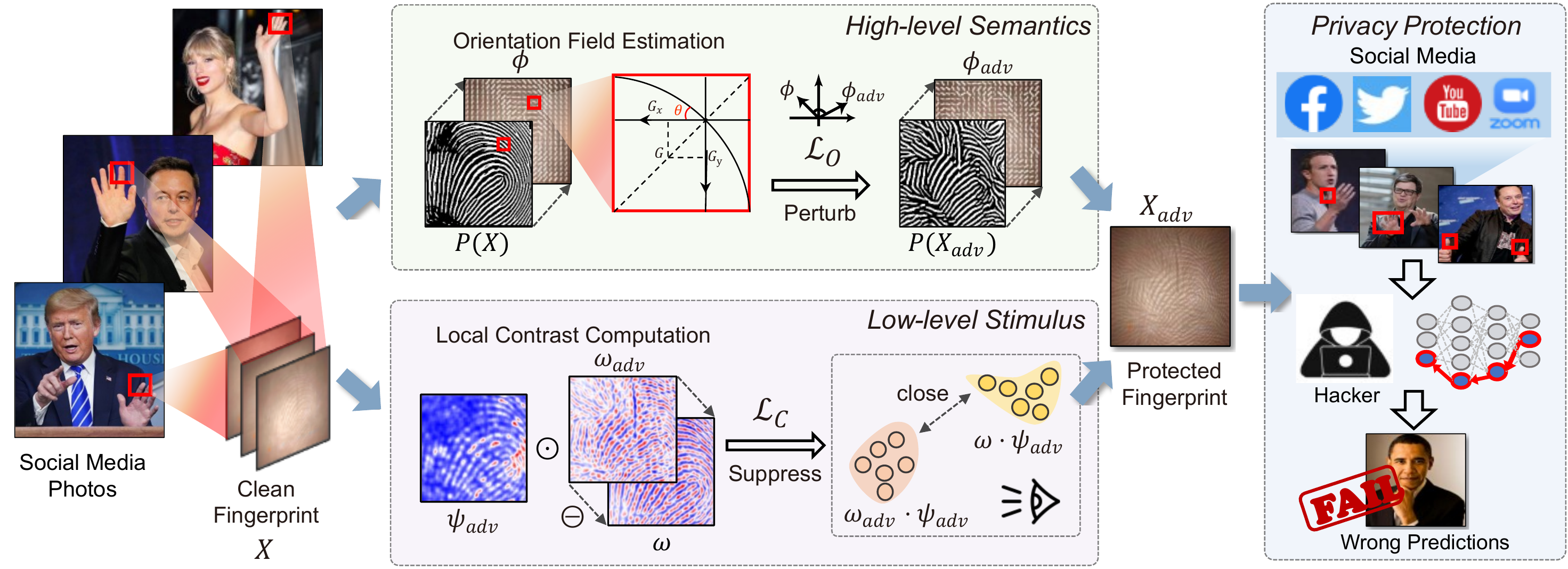}
\caption{Illustration of FingerSafe framework. Our hierarchical protective noise injection framework improves the transferability by perturbing high-level ridge semantics and achieves visual naturalness by suppressing low-level stimulus.}
\label{framework}
\end{figure*}

\subsection{FingerSafe Framework}

Our work is motivated by the visual perception hierarchy that different levels of perception focus on different abstract stages and contribute differently to model perception and human perception. As shown in Fig.\ref{framework}, our FingerSafe framework consists of two main modules, \ie, Orientation field distortion module and Local contrast suppression module, which respectively enforce strong protection transferability by attacking high-level semantic and enforce high naturalness by preserving low-level stimulus.

Regarding \emph{protection transferability}, high-level perception focus on predictions using abstract semantics that is shared between models. Similarly, perturbing semantics will be a model-agnostic attack that transfers well between different models. Specifically, we propose to estimate and distort fingerprint orientation field, a high-level feature of fingerprint. With distorted orientation field, the protected fingerprint is semantically different from the previous raw image, thus naturally transfer between different purposes, model architectures and preprocessing methods.

Regarding \emph{visual naturalness}, motivated by stimulus-response theory in psychology, the suspiciousness of human cognition is triggered by local contrast, the low-level stimuli. Clearly, naturalness will be enhanced if such low-level stimuli are not present. We first calculate the stimuli in human visual system, then add a regularization term to suppress the sensitive stimuli (local contrast) in FingerSafe, making protected images appropriate to be shared on social media.

\subsection{Orientation Field Distortion}
Previous works found the high-level features reflect the discriminative semantics towards a specific class, which are shared between models \cite{groen2017lowtohigh,simonyan2013deepinsideCNN,yosinski2015understandingCNN,wang2021dualattention}. Thus, attacking the model-shared high-level semantics could improve its transferability among models. As for fingerprint recognition, ridge feature is a high-level representation that reflects the uniqueness of fingerprint to determine identity \cite{minaee2019biometrics}, which is often considered lifetime invariant. Therefore, ridge features are often shared by different models, and recognition is built upon these preprocessed ridge features from clean fingerprints \cite{hong1998tpami,sankaran2015ijcb,lin2017ijcb}.

However, directly attacking ridge features are difficult since ridge features $\mathbf P(\mathbf x)$ is hidden in fingerprint image $\mathbf x$. Besides, the model (including preprocessing algorithms) used to extract ridge features are unknown to us. To reliably perturb ridge features, we inject protective noise on fingerprint orientation field, an estimate of the pixelwise orientation of ridge feature $\mathbf P(\mathbf x)$, which reflects the intrinsic trait of fingerprints and is agnostic to different models. With their close relationship, the distorted orientation field will force change in underlying ridge features. As a result, different models that rely on ridge features will fail to make correct predictions when false ridge features are given.

Specifically, we maximize the differences in orientation field between clean image $\mathbf x$ and protected image $\mathbf x_{adv}$ using least mean square orientation estimation \cite{hong1998tpami} with an orientation estimation module $\Phi$ as:
\begin{equation}
\label{eqn:getorientation}
\phi = \Phi \left (x \right ) = \frac{1}{2}\tan^{-1} \frac{\mathcal G\ast (2\mathcal G_{x}\ast \mathbf x \odot \mathcal G_{y} \ast \mathbf x)}{\mathcal G \ast \left ( (\mathcal G_{x} \ast \mathbf x)^{2}-(\mathcal G_{y}\ast \mathbf x )^2 \right )}+\frac{\pi}{2},
\end{equation}
where $\ast$ is convolution operator, $\odot$ is pointwise Hadamard product, $\mathcal G_{x}$ and $\mathcal G_{y}$ are Gaussian derivative kernels at direction $x$ and $y$, and $\mathcal G$ is Gaussian kernel to average the orientation estimates and avoid ambiguity.

Given the orientation field $\phi_{adv}$ calculated by Eqn. \ref{eqn:getorientation}, we introduce the orientation field distortion loss $\mathcal{L}_{O}$ as:
\begin{equation}
\label{eqn:orientationloss}
    \mathcal{L}_{O} = - \frac{1}{HW} \sum_{h}^{H} \sum_{w}^{W} \left \| \sin \left( \left|\phi_{adv}^{h, w} - \phi^{h, w} \right|\right)\right \|_{1},
\end{equation}
where $\| \cdot \|_{1}$ is the $\ell_1$ norm, H and W are the height and width of image $\mathbf x$. Since $\phi_{adv}$ and $\phi$ are angles, we use $\sin(\cdot)$ function to measure their difference.

Additionally, we adopt an adversarial loss $\mathcal L_{adv}$ to further guide the protective nohhise generation process. $\mathcal L_{adv}$ maximizes the distance of the representation calculated from our source model $\mathbb F_{\vartheta}$ between protected image and clean image as
\begin{equation}
\begin{aligned}
\
\label{eqn:advloss}
\mathcal{L}_{adv} &= - \|\mathbb F_{\vartheta}(\mathbf x_{adv}) - \mathbb F_{\vartheta}(\mathbf x) \|_{2}. \\
\end{aligned}
\end{equation}

\subsection{Local Contrast Suppression}
To achieve reliable privacy protection, existing works usually generate attacks with large perturbation budgets \cite{yang2020JunZhu,cherepanova2021lowkey}, which cause suspicious appearance and reduced utility. Motivated by extensive biological observations that human vision is highly suspicious towards the  subtle variation of low-level stimulus (\ie, local contrast), we thereof generate protected images with higher naturalness by protecting these sensitive features intact.

Specifically, local contrast perceived by human are characterized by ganglion cells of the retina and the Lateral Geniculate Nucleus (LGN) neurons, and their responses could be calculated by a modified differences of Gaussian (DOG) model \cite{tadmor2000localcontrast}. We therefore introduce a local contrast calculation module $\Omega$ as:
\begin{equation}
\begin{aligned}
\label{eqn:contrast}
\mathcal G_{c}(i, j) &= \exp \left[ - \left(i/r_c \right)^2 - \left(j/r_c\right)^2 \right],\\
\mathcal G_{s}(i, j) &= 0.85 \left(r_c/r_s\right)^2\exp \left[ -\left(i/r_s\right)^2 - \left(j/r_s\right)^2 \right],\\
\omega &= \Omega(\mathbf x) = \frac{\mathcal G_c \ast \mathbf x - \mathcal G_s \ast \mathbf x}{\mathcal G_c \ast \mathbf x + \mathcal G_s \ast \mathbf x}, \\
\end{aligned}
\end{equation}
where $\mathcal G_{c}$ and $\mathcal G_{s}$ are two Gaussian kernels that calculates the center and surrounding component with receptive field $r_c$ and $r_s$. $\Omega(\mathbf x)$ is simply a DOG model with the term $\mathcal G_c \ast \mathbf x + \mathcal G_s \ast \mathbf x$ to resemble the light adaptation process of ganglion cells of the retina and LGN.

However, simply keeping local contrast intact shares the opposite objective with fingerprint orientation field distortion, which may fail to provide protection. Thus, to release the constraint of regularizing local contrast on all parts of the image, we propose a local contrast attention module $\Phi$ to identify the importance of different regions within an image \cite{veale2017saliencebrain} and thereby inject different perturbation budgets, which could be written as:
\begin{equation}
\label{eqn:saliency}
\psi = \Psi(\mathbf x)=\mathcal G \ast \mathcal{F}^{-1} \left[ \exp(\mathcal I - \mathcal B\ast \mathcal I) + \mathcal{P} \right]^2,
\end{equation}
where $\mathcal{A}= Re(\mathcal{F}(\mathbf x))$, $\mathcal{P}= Im(\mathcal{F}(\mathbf x))$ and $\mathcal{I}= \log(\mathcal{A})$ are the amplitude spectrum, phase spectrum and logarithm of amplitude spectrum. $\mathcal{F}$ and $\mathcal{F}^{-1}$ are Fourier and inverse Fourier transform.  $\mathcal G$ and $\mathcal B$ are Gaussian kernel and box kernel, respectively. Note that $\psi_{adv}$ was used to identify important areas in both clean and protected images, H and W are the height and width of image $\mathbf x$. Thus, the local contrast suppression loss $\mathcal{L}_{C}$ can be written as:
\begin{equation}
\label{eqn:humanperceptionloss}
    \mathcal{L}_{C} = \frac{1}{HW} \sum_{h}^{H} \sum_{w}^{W} ReLU\left(\left(\omega_{adv}^{h, w} - \omega^{h, w}\right) \odot \psi_{adv}^{h, w} \right).
\end{equation}

\begin{algorithm}[t]
\caption{FingerSafe Privacy Protection}
\label{alg1}
\begin{algorithmic}[1]
\renewcommand{\algorithmicrequire}{\textbf{Input:}}
\renewcommand{\algorithmicensure}{\textbf{Output:}}
\REQUIRE Fingerprint image database $\mathcal{D}$, orientation estimation module $\Phi$, local contrast estimation module $\Omega$, saliency map estimation module $\Psi$.
\ENSURE Protected image database $\mathcal{D}_{adv}$
\FOR{Minibatch $\mathbf x$ in dataset $\mathcal{D}$}
\STATE $\mathbf x_{adv}^{0} = \mathbf x$, $i=0$
\FOR{i in number of iterations}
\STATE $\phi \gets \Phi(\mathbf x)$, $\phi_{adv}^{i} \gets \Phi(\mathbf x_{adv}^{i})$ by Eqn.\ref{eqn:getorientation}
\STATE $\omega \gets \Omega(\mathbf x)$, $\omega_{adv} \gets \Omega(\mathbf x_{adv})$ by Eqn. \ref{eqn:contrast}
\STATE $\psi \gets \Psi(\mathbf x)$, $\psi_{adv} \gets \Psi(\mathbf x_{adv})$ by Eqn. \ref{eqn:saliency}
\STATE calculate $\mathcal{L}_{O}$, $\mathcal{L}_{C}$ and $\mathcal{L}_{adv}$ by Eqn. \ref{eqn:orientationloss}, \ref{eqn:humanperceptionloss}, \ref{eqn:advloss}.
\STATE $\mathbf x_{adv}^{i+1} \gets \mathbf x_{adv}^{i}+\alpha sgn (\nabla_{\mathbf x_{adv}^{i}}(\mathcal{L}_{adv}+\lambda \mathcal{L}_{O} +\gamma \mathcal{L}_{C}))$ 
\ENDFOR
\STATE $\mathcal{D}_{adv} \gets \mathbf x_{adv}^{i+1}$
\ENDFOR
\end{algorithmic}
\end{algorithm}

\subsection{Overall Training}
Overall, we generate the protected image by jointly optimizing the adversarial loss $\mathcal{L}_{adv}$, orientation field distortion loss $\mathcal{L}_{O}$, and local contrast suppression loss $\mathcal{L}_{C}$. The implementation of $\mathcal{L}_{O}$ and $\mathcal{L}_{C}$ includes only Gaussian smoothing and Fast Fourier Transform. As a result, the execution time of FingerSafe is comparatively similar with state-of-the-art adversarial attacks \cite{madry2017pgd}. Using iterative gradient descent based method \cite{madry2017pgd}, we can generate transferable and visually natural protected image $\mathbf x_{adv}$ by minimizing the formulation below:


\begin{equation}
\label{overallopt}
\min_{\mathbf x_{adv}} \ \mathcal{L}_{adv} + \lambda \mathcal{L}_{O} + \gamma \mathcal{L}_{C},
\end{equation}
where $\lambda$ and $\gamma$ are hyperparameters to control the strength of $\mathcal{L}_{O}$ and $\mathcal{L}_{C}$, respectively. To balance attacking ability $\mathcal{L}_{adv}$, transfer ability $\mathcal{L}_{O}$ and visual naturalness $\mathcal{L}_{C}$, we set $\lambda$ as $10^{2}$ and $\gamma$ as $5 \times 10^{2}$, where small $\mathcal{L}_{O}$ cannot perturb high-level semantics sufficiently and high $\mathcal{L}_{C}$ will wipe out all the protection pattern. The overall training algorithm for FingerSafe can be described in Algorithm. \ref{alg1}. 

\section{Experiments}
In this section, we first describe our experiment settings, then evaluate the transferability, naturalness and real-world social media protection of FingerSafe.

\subsection{Experimental Setup}
\subsubsection{\textbf{Dataset and target models}}
We conduct our experiment on HKPolyU Database \cite{lin2018HKPolyU}, a widely used fingerprint dataset with photoed fingerprints from 336 different subjects. While most images in social media are colored, the given dataset is in grayscale. We use an off-the-shelf image colorization method \cite{lu2020gray2colornet} to transform the whole dataset into RGB format. 

Regarding fingerprint recognition, as for the deep learning based methods, we follow the setting of \cite{malhotra2020cvprw,chopra2018unconstraineddatabase}, which first extracts fingerprint features by a DNN model, then pair the images using Euclidean distance. We consider two tasks for fingerprint recognition based on the output including verification (determine if a pair of fingerprints belongs to the same image) and identification (find the identity with given fingerprint). For DNNs, we use ResNet50 \cite{2016Deep}, Inceptionv3 \cite{szegedy2016inception} and DenseNet121 \cite{huang2017densenet}. For traditional methods, we use ScatNet \cite{bruna2013scatnet} and minutiae matching \cite{wikeclaw2009minutiae}, where no parameters are learned. The preprocessing methods we use include MHS \cite{sankaran2015ijcb}, HG \cite{lin2017ijcb} and Frangi filter \cite{frangi1998frangi} \footnote{We do not test on DNN-based minutiae extraction method since they need the ground-truth minutiae. HKPolyU dataset does not provide such data}.

\subsubsection{\textbf{Compared methods and evaluation metrics}}
For attack methods that protect privacy, we choose the current state-of-the-art methods including PGD \cite{madry2017pgd}, Unlearnable example (denoted Unlearn.) \cite{ICLR2021unlearnable}, Fawkes \cite{shan2020fawkes}, LowKey\cite{cherepanova2021lowkey} and Malhotra et, al. \cite{malhotra2020cvprw} (denoted as Malhotra). Among these baselines, Unlearnable examples are designed for training stage protection, while others aim at testing stage protection\footnote{We use their released code and tuned their hyperparameters for best performance.}. To better evaluate the performance, we use model accuracy (ACC) as the evaluation metric for identification. For verification, we use TPR as evaluation metric, since using accuracy will confuse true positive rate (TPR) and true negative rate (TNR), where an ideal verification protection method should keep TPR=0\% and TNR=100\%, (\ie, hackers cannot sneak in the FingerID system of identity $i$ using image $\mathbf x_{adv}^i$, while the ground truth fingerprint stored in third-party system is $\mathbf x_i$). Such ideal protection yields the best verification accuracy of 50\%, but remains ambiguous since another algorithm with the same verification accuracy could have TPR=100\% and TNR=0\% (\ie, any hacker could sneak in the fingerprint verification system without failure). 

In particular, in testing stage protection, we assume hackers collect fingerprints from internet, and tries to sneak in well-trained fingerprint recognition system, with a clean fingerprint database. As a result, we train the recognition model using clean image pairs, then use the clean-protected image pairs to evaluate model accuracy. In training stage protection, we assume hackers collect additional fingerprints from website to enhance the performance of their FingerID system, like \emph{clearview.ai} \cite{hill2020clearview}. Thus, we train the recognition model using protected image pairs, then use clean image pairs to verify model performance.

\subsubsection{\textbf{Implementation details}}
In orientation estimation module $\Phi$, $\mathcal G_x$ and $\mathcal G_y$ are 7*7 derivatives of Gaussian kernel on x and y direction. Gaussian kernel $\mathcal G$ has size 31*31. In local contrast calculation module $\Omega$, $r_c$ and $r_s$ is set to 2 and 4, the size of $G_c$ and $G_s$ is 13*13 and 25*25, respectively. In saliency map calculation module $\Psi$, $\mathcal G$ is a Gaussian kernel with size 9*9, $\mathcal B$ is a box kernel with size 3*3. These kernel sizes and parameters follow the original implementation. For compared methods, we set dissimilarity threshold to 0.1 to generate Fawkes adversarial samples, and parameter of LPIPS to 5 to generate LowKey samples. When extracting ScatNet feature, the number of scales $J$ and orientations $L$ is set to 2 and 8, respectively, with input shape of (50, 50).

We empirically set $\gamma = 10^{2}$, $\lambda = 5 \times 10^{2}$ for protected image generation. The maximum perturbation step is 20. All learnable parameters are optimized by an Adam optimizer with a learning rate of $10^{-4}$ and a maximum of 50 epochs. We keep all the methods for the same magnitude of perturbations $\epsilon$=8/255 in terms of $\ell_{\infty}$-norm. All of our codes are implemented in PyTorch \cite{paszke2017pytorch}. We conduct all the experiments on a NVIDIA Tesla V100-SXM2-16GB GPU cluster.

For all of our experiments, the fingerprint images are either obtained from public HKPolyU dataset with a license agreement, or collected from participants who signed an agreement with us under supervision of ethic commitee to assure that their information will be used in non-commercial research.

\subsection{Transferability of FingerSafe}
In this section, we conduct extensive experiments to evaluate the transferability of FingerSafe in black-box settings on multiple model architectures, preprocessors, traditional non-learning methods. We show the performance of different purposes (verification and identification, denoted as Veri. and Iden.) in each case. 

\begin{table*}[]
\caption{Transfer to different architectures. We use ResNet50, InceptionV3, DenseNet121 and an ensemble of three models as source model and transfer to different target models. FingerSafe outperforms baselines by perturbing ridge features shared by different models.}
\label{digital-architecture}
\centering
\setlength\tabcolsep{10pt}
\begin{tabular}{@{}cccccccc@{}}
\toprule
\multirow{2}{*}{From/To}     & Task       & \multicolumn{3}{c}{Veri. TPR(\%)}             & \multicolumn{3}{c}{Iden. ACC(\%)}             \\ \cmidrule(l){2-8} 
                             & Method     & Resnet50      & InceptionV3   & DenseNet121   & ResNet50      & InceptionV3   & DenseNet121   \\ \midrule
-                            & Clean      & 91.17         & 93.28         & 96.39         & 94.12         & 91.18         & 94.85         \\ \midrule
\multirow{5}{*}{ResNet50}    & PGD        & 3.48          & 43.28         & 3.36          & 10.29         & 33.82         & 12.50         \\
                             & Unlearn.   & 5.58          & 54.73         & 5.85          & 16.18         & 41.18         & 18.38         \\
                             & Fawkes     & 0.00          & 15.30         & 2.11          & \textbf{0.00} & 10.29         & 7.35          \\
                             & Malhotra   & 5.35          & 50.50         & 13.31         & 12.50         & 46.32         & 35.29         \\
                             & FingerSafe & \textbf{0.00} & \textbf{3.86} & \textbf{1.62} & \textbf{0.00} & \textbf{5.88} & \textbf{2.21} \\ \midrule
\multirow{5}{*}{InceptionV3} & PGD        & 24.75         & 12.19         & 23.76         & 28.67         & 20.59         & 30.15         \\
                             & Unlearn.   & 1.74          & 11.07         & 2.99          & 8.09          & 5.15          & 7.35          \\
                             & Fawkes     & 0.50          & 15.30         & 4.60          & 6.62          & 3.23          & 6.62          \\
                             & Malhotra   & 1.99          & 14.47          & 22.61          & 7.35          & 24.26          & 29.41          \\
                             & FingerSafe & \textbf{0.12} & \textbf{3.23} & \textbf{0.37} & \textbf{3.68} & \textbf{2.21} & \textbf{2.21} \\ \midrule
\multirow{5}{*}{DenseNet121} & PGD        & 10.32         & 57.96         & 0.25          & 15.44         & 44.85         & 5.88          \\
                             & Unlearn.   & 2.61          & 31.09         & 0.12          & 5.15          & 4.41          & 2.94          \\
                             & Fawkes     & 0.37          & 14.55         & \textbf{0.00} & 6.62          & 9.56          & 1.47          \\
                             & Malhotra   & 2.49          & 19.03          & 2.11          & 2.94          & 5.88          & 2.21          \\
                             & FingerSafe & \textbf{0.00} & \textbf{4.73} & \textbf{0.00} & \textbf{2.94} & \textbf{2.21} & \textbf{0.00} \\ \midrule
Ensemble                     & LowKey     & 0.87          & 11.57         & 3.23          & 3.68          & 7.35          & 3.70          \\ \bottomrule
\end{tabular}
\end{table*}

\subsubsection{\textbf{Different deep learning architectures}}
We start from the case that hackers use black-box DNN models for recognition, and FingerSafe aims to generate images that are able to transfer between different black-box architectures. For fair comparison, we keep other settings the same for different architectures. Here, we use ResNet50, InceptionV3, DenseNet121 and an ensemble of these models (used in Lowkey) as our source model to generate adversarial examples, and use ResNet50, InceptionV3 and Densenet121 as the target models (the case that using the same source model and target model indicates the white-box attacks otherwise indicates black-box attacks. As listed in Tab.\ref{digital-architecture}, we can draw several conclusions as follows:

(1) For black-box attacks, FingerSafe consistently demonstrates superior transferability and outperforms LowKey, the best performing baselines by large margins (\ie, up to \textbf{7.71\%} protection improvement over best performing method).

(2) For white-box attacks (ResNet50$\rightarrow$ResNet50, InceptionV3$\rightarrow$InceptionV3, DenseNet121$\rightarrow$DenseNet121), FingerSafe also consistently provides stronger protection, outperforming Unlearn. by large margins in InceptionV3$\rightarrow$InceptionV3 (\ie, up to \textbf{7.84\%} protection improvement over best performing method).

\begin{figure*}[t]
\centering
\includegraphics[scale=0.5]{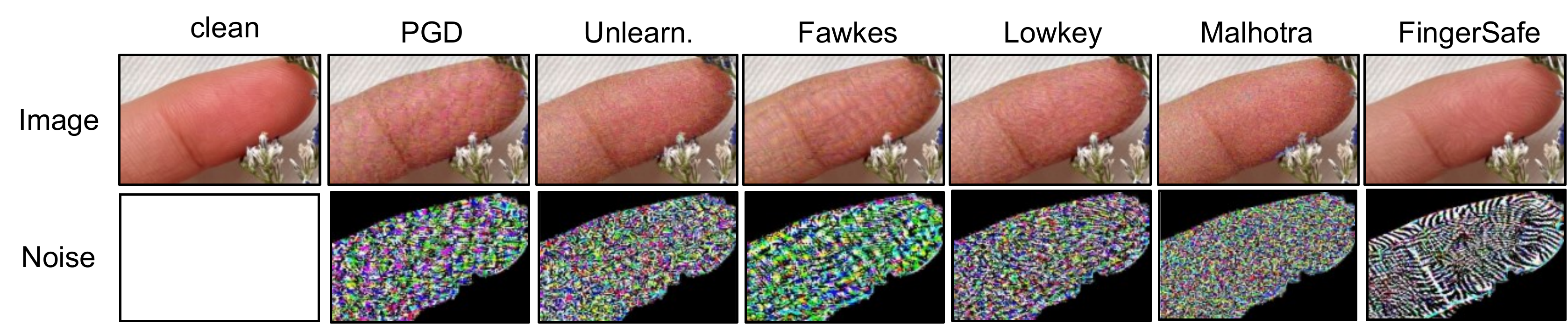}
\caption{Visualization of protected images and protective noises under same protection capability. The noise of FingerSafe shows fingerprint-like high-level semantic, achives stronger transferability under various transfer settings.}
\label{visualizenoise}
\end{figure*}


(3) Comparing with FingerSafe, baselines hardly transfer well to different models, especially InceptionV3. To understand it, regarding the noises in Fig.\ref{visualizenoise}, the noises generated by baselines do not contain semantic-related component, while the protective noise of FingerSafe shows fingerprint-alike high-level semantics, which are shared by different DNN architectures. Thus, protected images generated by FingerSafe are likely to be viewed as a \emph{semantically different} fingerprint, which is model-agnostic.



\subsubsection{\textbf{Different preprocessors}}

We further assume hackers use unknown preprocessors to extract ridge features and use them for further recognition, which is common in fingerprint processing \cite{jain1997tpamiminutiae,frangi1998frangi,sankaran2015ijcb,lin2017ijcb}. To further validate the effectiveness of our FingerSafe in such a more practical scenario, we generate protected images on a ResNet50 source model without preprocessors, and then transfer to a target ResNet50 model which uses a variety of preprocessors including MHS, HG, and Frangi filter. As shown in Tab. \ref{digital-preprocessor}, FingerSafe achieves significantly strong performance and outperforms others by large margins when preprocessors were used for target models (\ie, up to \textbf{68.91\%} protection improvement over best performing method). Importantly, baseline methods are ineffective in this scenario.

\begin{figure}[t]
\centering
\includegraphics[scale=0.27]{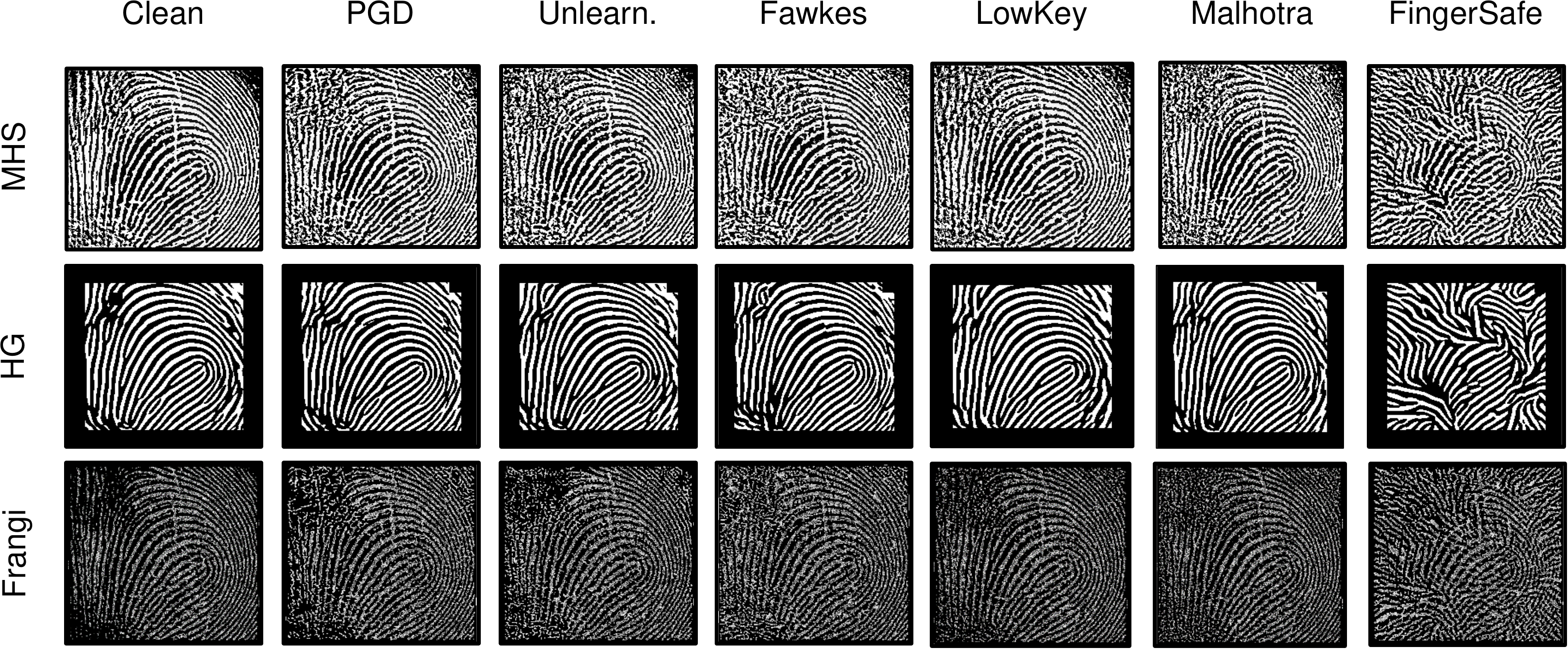}
\caption{Visualization of preprocessed image under same $\ell_\infty$ constraint. By perturbing high-level semantics, FingerSafe fundamentally changes the preprocessed fingerprint.}
\label{preprocess}
\end{figure}

To better understand the performance of FingerSafe in this setting, we visualize the fingerprint extracted by all preprocessing methods in Fig. \ref{preprocess}. As we have mentioned in section 3.3, the goal of preprocessing is to extract high-level ridge features with rich semantics, while discarding low-level features. As a consequence, the noise generated by baselines contains both high-level ridge features and low-level non-ridge features. After preprocessing, all
non-ridge noises are simply filtered out (i.e., baselines failed to attack high-level semantic). In contrast, the protective noise of FingerSafe shows high-level, ridge-like patterns that could attack different preprocessors. Thus, hackers cannot even extract the true fingerprint from images protected by FingerSafe, let alone use it for correct recognition.

\begin{table}[t]
\caption{Experiment results for different preprocessors. FingerSafe significantly outperforms all baselines by perturbing ridge features.}
\label{digital-preprocessor}
\centering
\setlength\tabcolsep{2pt}
\scriptsize
\begin{tabular}{@{}ccccccccc@{}}
\toprule
Metric                                                                   & Preprocessor & Clean & PGD   & Unlearn. & Fawkes & Lowkey & Malhotra       & \textbf{FingerSafe}    \\ \midrule
\multirow{3}{*}{\begin{tabular}[c]{@{}c@{}}Veri.\\ TPR(\%)\end{tabular}} & MHS          & 93.03 & 68.28 & 67.54    & 69.43  & 77.74  & 59.33                          & \textbf{5.85} \\ \cmidrule(l){2-9} 
                                                                         & HG           & 88.06 & 82.90 & 82.09    & 73.51  & 80.72  & 81.09                          & \textbf{4.60} \\ \cmidrule(l){2-9} 
                                                                         & Frangi       & 89.43 & 22.01 & 24.88    & 7.34   & 51.99  & 32.09                          & \textbf{6.59} \\ \midrule
\multirow{3}{*}{\begin{tabular}[c]{@{}c@{}}Iden.\\ ACC(\%)\end{tabular}} & MHS          & 93.38 & 49.26 & 48.89    & 41.48  & 40.44  & 25.74                          & \textbf{0.74} \\ \cmidrule(l){2-9} 
                                                                         & HG           & 72.79 & 61.76 & 60.29    & 51.47  & 58.09  & 54.41                          & \textbf{0.00} \\ \cmidrule(l){2-9} 
                                                                         & Frangi       & 93.38 & 13.97 & 16.91    & 5.15   & 6.62   & 11.76                           & \textbf{1.47} \\ \bottomrule
\end{tabular}
\end{table}

\subsubsection{\textbf{Traditional non-learning methods}}
\begin{figure}[t]
\centering
\includegraphics[scale=0.27]{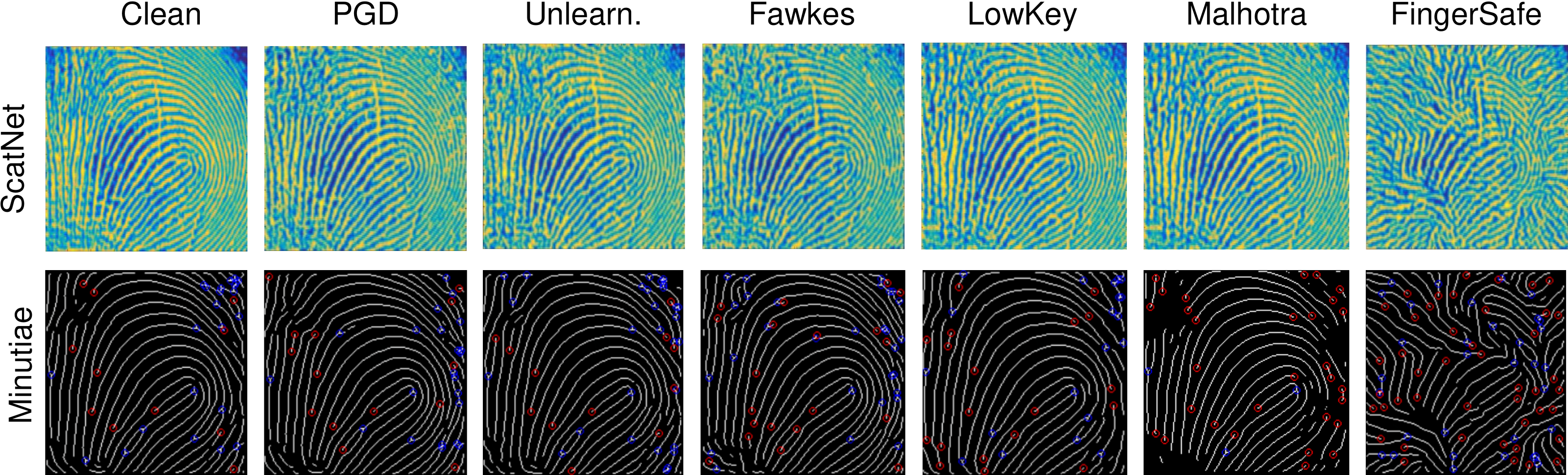}
\caption{Visualization of ScatNet features and minutiae under different protection methods. By perturbing high-level ridge semantics, FingerSafe is the only method to change the traditional features fundamentally.}
\label{nonlearning}
\end{figure}

In this section, we assume hackers use the traditional non-learning methods where no parameters were learned. In particular, we generate protected images on a ResNet50 source model and transfer it to Scattering Network (ScatNet)\cite{malhotra2020scatnet} and minutiae-based matching \cite{wikeclaw2009minutiae}. Specifically, we use $\ell_2$ distance as similarity between two ScatNet features. For minutiae matching, we extract minutiae by \cite{hong1998tpami,wikeclaw2009minutiae} and use the method described in \cite{wikeclaw2009minutiae} to match paired minutiae. As shown in Tab.\ref{digital-Scat}, FingerSafe achieves significantly strong protection performance on black-box traditional non-learning models, which improves protection capability over best baseline by large margins (up to \textbf{63.28\%} in verification and \textbf{50.73\%} in identification), demonstrating strong transferability even no parameters are learned. We also provide an illustration of the extracted ScatNet feature and minutiaes in Fig. \ref{nonlearning}. We show that by changing high-level ridge semantics, FingerSafe can also fundamentally perturb the extracted traditional features even when no parameters were learned in overall fingerprint recognition process.

\begin{table}[h]
\caption{Experiment results for non-learning methods. FingerSafe is effective even no parameters are learned.}
\label{digital-Scat}
\centering
\setlength\tabcolsep{2pt}
\scriptsize
\begin{tabular}{@{}ccccccccc@{}}
\toprule
Metric                                                                   & Method         & Clean & PGD   & Unlearn. & Fawkes & Lowkey & Malhotra    & \textbf{FingerSafe}     \\ \midrule
\multirow{2}{*}{\begin{tabular}[c]{@{}c@{}}Veri.\\ TPR(\%)\end{tabular}} & MHS-Scat-L2    & 76.37 & 72.14 & 74.75   & 57.09   & 70.77 & 67.08                        & \textbf{9.70} \\ \cmidrule(l){2-9} 
                                                                         & MHS-Minu. & 89.63 & 70.73 & 71.95   & 68.29        & 73.45 & 68.16             & \textbf{4.88} \\ \midrule
\multirow{2}{*}{\begin{tabular}[c]{@{}c@{}}Iden.\\ ACC(\%)\end{tabular}} & MHS-Scat-L2    & 83.82 & 66.18 & 70.59   & 39.71   & 61.03 & 33.82                       & \textbf{3.68}  \\ \cmidrule(l){2-9} 
                                                                         & MHS-Minu. & 85.29 & 69.85 & 68.38   & 61.76        & 65.44 & 55.88             & \textbf{5.15}  \\ \bottomrule
\end{tabular}
\end{table}

\begin{table}[h]
\caption{Experiment result for training stage protection. Tested on both different architectures and different preprocessing methods. Note that ResNet50 corresponds to white-box attack. Despite being designed for testing stage protection, FingerSafe also outperforms all baselines in training stage protection.}
\label{digital-training}
\centering
\setlength\tabcolsep{1pt}
\scriptsize
\begin{tabular}{@{}ccccccccc@{}}
\toprule
Metric                                                                   & Method       & Clean & PGD   & Unlearn. & Fawkes & Lowkey & Malhotra   & \textbf{FingerSafe}     \\ \midrule
\multirow{4}{*}{\begin{tabular}[c]{@{}c@{}}Veri.\\ TPR(\%)\end{tabular}} & ResNet50     & 91.17 & 92.54 & 79.98    & 86.94  & 81.97  & 89.05                      & \textbf{79.10} \\ \cmidrule(l){2-9} 
                                                                         & InceptionV3  & 93.28 & 92.41 & 90.55    & 89.68  & 94.78  & 92.29                      & \textbf{85.32} \\ \cmidrule(l){2-9} 
                                                                         & MHS-ResNet50 & 95.14 & 90.67 & 91.79    & 89.30  & 91.92  & 92.16                      & \textbf{83.33} \\ \cmidrule(l){2-9} 
                                                                         & HG-ResNet50  & 88.06 & 89.93 & 85.95    & 84.70  & 87.81  & 86.32                      & \textbf{80.97} \\ \midrule
\multirow{4}{*}{\begin{tabular}[c]{@{}c@{}}Iden.\\ ACC(\%)\end{tabular}} & ResNet50     & 94.12 & 86.03 & 84.56    & 83.88  & 81.62  & 82.35                      & \textbf{75.74} \\ \cmidrule(l){2-9} 
                                                                         & InceptionV3  & 91.18 & 90.44 & 90.44    & 89.71  & 90.44  & 92.65                      & \textbf{83.82} \\ \cmidrule(l){2-9} 
                                                                         & MHS-ResNet50 & 93.38 & 86.76 & 86.03    & 91.17  & 86.03  & 91.91                      & \textbf{80.88} \\ \cmidrule(l){2-9} 
                                                                         & HG-ResNet50  & 72.79 & 69.18 & 71.32    & 62.50  & 64.71  & 70.59                      & \textbf{47.79} \\ \bottomrule
\end{tabular}
\end{table}

\subsubsection{\textbf{Training stage protection}}
In this part, we demonstrate that FingerSafe can be directly used to protect \emph{training stage attack} without any modification. In contrast to testing stage protection, \emph{training stage protection} aims to release protected image $\mathbf x_{adv}$ on social media, so hacker train their model $\mathbb F_{\Theta}$ on $\mathbf x_{adv}$ yields low performance when testing on clean images $\mathbf x$. 
In particular, we generate all protected images based on ResNet50 source model and then feed these images to hackers to train their models (different architectures and different preprocessing methods)\footnote{Result on traditional non-learning methods were not shown since no training is required.}. The results in Tab.\ref{digital-training} lead to several conclusions:

(1) In training stage protection, FingerSafe achieves best performance in all black-box settings by large margins (\ie, up to \textbf{14.71\%} protection improvement over the best baseline). Our method even outperforms Unlearnable examples by large margins which is specifically designed for training stage protection.

(2) FingerSafe is very effective in white-box settings (ResNet50$\rightarrow$ResNet50), leading a \textbf{8.14\%} protection improvement over the best baseline. 

We hypothesis the better performance of FingerSafe in training stage protection is that our method aims to attack the high-level semantics (\ie, ridge patterns). These features are key evidences for models to learn and perform correct predictions. Thus, perturbing high-level semantics could be interpreted as adding a robust ``watermark'' that provides stable feature collision, which is harder for DNN to separate at training time \cite{shafahi2018poisonfrog}.

\subsection{Naturalness of FingerSafe}

\begin{figure}[h]
\centering
\includegraphics[scale=0.12]{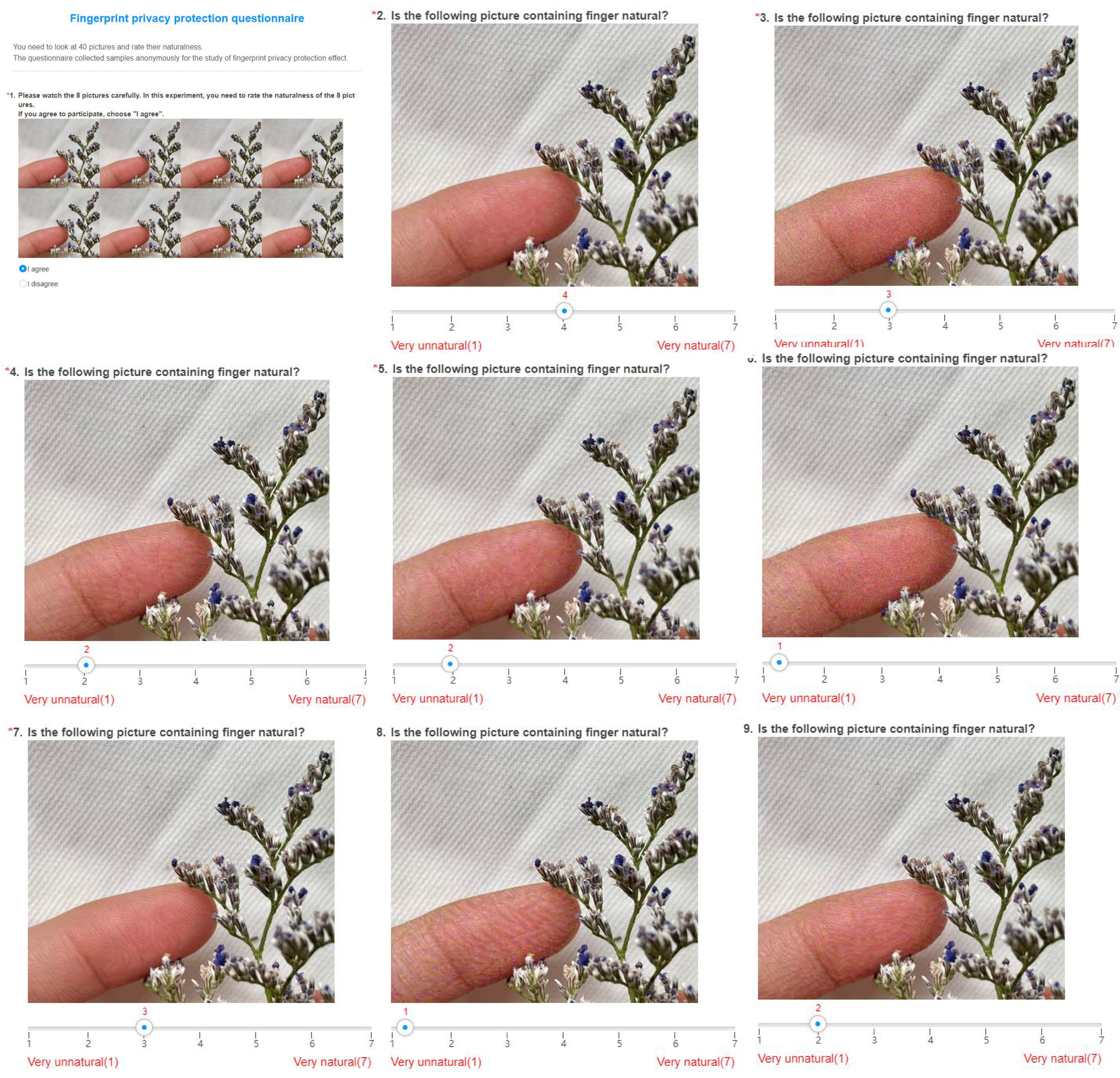}
\caption{Questionnaire on the naturalness of finger images issued to users. The participants are needed to rate the naturalness of images containing fingers.}
\label{question}
\end{figure}

\begin{figure}[h]
\centering
\includegraphics[width=0.9\linewidth]{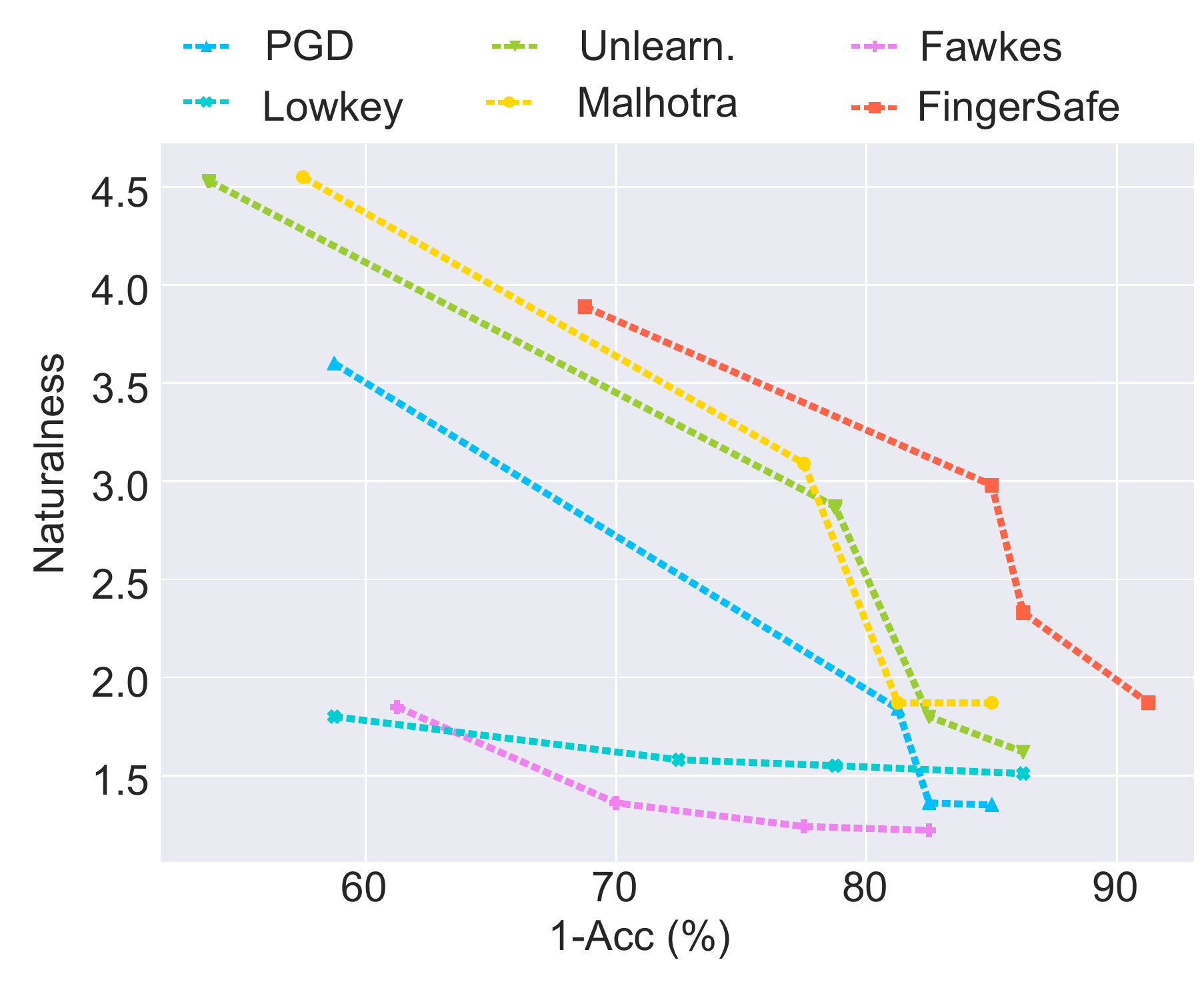}
\caption{Human perception studies on the naturalness of different methods. FingerSafe achieves better performance on both protection capability and naturalness.}
\label{humanexp}
\end{figure}

The quality of the fingerprint privacy protection also depends on the naturalness of protected images. We therefore conduct human perception studies on one of the most commonly used crowdsourcing platforms. In particular, participants were asked to evaluate the naturalness of social media images protected by PGD, Unlearnable examples, Fawkes, Lowkey, Malhotra and FingerSafe. Since naturalness cannot be fairly evaluated without protection performance, we select four levels of protection strength using four different $\ell_{\infty}$ constraints. All ratings were collected by a 7-point Likert table, from 1 (very low) to 7 (very high). We collect all responses from 55 anonymous participants, with a sample of our questionnaire shown in Fig. \ref{question}. The result was illustrated in Fig.\ref{humanexp}, where Acc refers to identification accuracy. The goal is to achieve better protection capability (we use 1 minus Acc for convenience) and higher naturalness.

(1) FingerSafe simultaneously provides better results in both protection capability and naturalness than all baselines. It is worth noting that our Fingersafe performs especially better than other baselines by large margins under high perturbation budget, which show high application potentials in practice.

(2) FingerSafe is the only method to provide comparatively natural performance (2.98) under satisfying protection capability (\textgreater 85\%), achieves \textbf{59.36\%} improvement compared with Malhotra, the best performing baseline (2.98 vs 1.87). 

(3) State-of-the-art face privacy protection methods, such as Fawkes and Lowkey, is of low naturalness regardless of the $\ell_\infty$ constraint. Those methods cannot be used directly for fingerprint privacy protection.

\subsection{Social media protection: a real-world study}
\begin{figure}[t]
\centering
\includegraphics[scale=0.32]{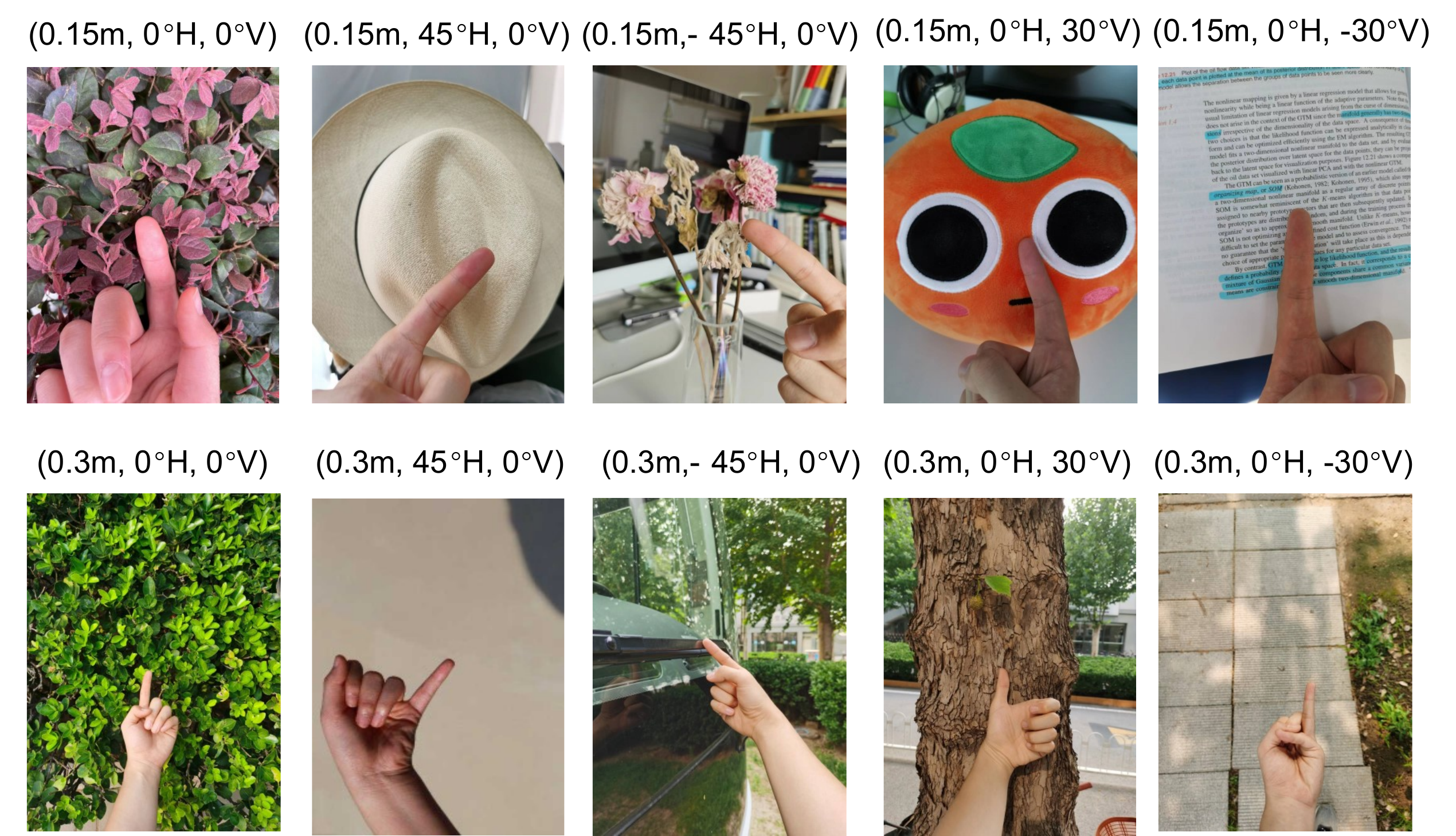}
\caption{Sample of collected data. We simulate images from different background, distance and angles.}
\label{dataset}
\end{figure}

In this section, we further test our FingerSafe in the real-world social media scenario. Specifically, we manually collect the fingerprint from 100 classes (identities) and obtain an overall 1000 images (each class contains 10 images) \footnote{All subjects signed an agreement before collecting their fingerprints. Experiment was conducted under supervision of ethics commitee.}. For each class, we took 10 photos with diversified background to simulate social media image, with different angles $\{-45^{\circ}, 0^{\circ}, 45^{\circ}\}$ horizontally, $\{-30^{\circ}, 30^{\circ}\}$ vertically, and different distances $\{0.15m, 0.3m\}$ using an iPhone 12 Pro camera. Sample images could be seen in Fig. \ref{dataset}. For each clean image, we first use FingerSafe to generate the protected image based on a ResNet50 source model and share them under private mode on Twitter and Facebook; we then download all images and provide an appropriate segmentation \cite{priesnitz2021pipeline,sankaran2015ijcb} for subsequent fingerprint recognition. Specifically, (1) after the image was acquired, we preprocess it to contain the fingerprint area only. (2) Following the methods of \cite{priesnitz2021pipeline,sankaran2015ijcb}, we first generate binary mask through erosion and dilation, find the largest contour, and then crop finger to the first knuckle. (3) We then use preprocessing algorithm to extract the ridge features. Note that noises are injected into masked area only. After noise injection, we record the adversarial noise and add it back to original position of the first knuckle. In this way, we could simulate the real-world hacking and protection processes, in which unknown models, preprocessors, \etc. are used in these platforms.

\begin{table}[h]
\caption{Results of FingerSafe in social media protection. We upload all images to social media (\ie, Twitter and Facebook), then download it to simulate real-world hacking. FingerSafe is effective in real-world social media protection.}
\label{physical-Twitter}
\centering
\setlength\tabcolsep{1pt}
\scriptsize
\begin{tabular}{@{}ccccccccc@{}}
\toprule
\multirow{2}{*}{Platform} & \multirow{2}{*}{Method} & \multicolumn{7}{c}{Iden. ACC (\%)}                  \\ \cmidrule(l){3-9} 
                          &                         & Clean & PGD   & Unlearn. & Fawkes & Lowkey & Malhotra  & \textbf{FingerSafe}     \\ \midrule
\multirow{3}{*}{Twitter}   & ResNet50               & 91.25 & 47.50 & 67.50    & 60.00  & 55.00  & 60.00 & \textbf{41.25} \\ \cmidrule(l){2-9} 
                          & MHS-ResNet50            & 76.25 & 42.50 & 47.50    & 31.25  & 41.25  & 41.25 & \textbf{15.00} \\ \cmidrule(l){2-9} 
                          & MHS-Scat-L2             & 78.75 & 61.25 & 65.00    & 45.00  & 57.50  & 56.25 & \textbf{37.50} \\ \midrule
\multirow{3}{*}{Facebook} & ResNet50                & 93.75 & 45.00 & 47.50    & 43.75  & 40.00  & 45.00 & \textbf{36.25} \\ \cmidrule(l){2-9} 
                          & MHS-ResNet50            & 80.00 & 40.00 & 46.25    & 38.75  & 41.25  & 38.75 & \textbf{11.25} \\ \cmidrule(l){2-9} 
                          & MHS-Scat-L2             & 88.75 & 77.50 & 83.75    & 75.00  & 76.50  & 78.75 & \textbf{52.50} \\ \bottomrule
\end{tabular}
\end{table}


The evaluation results can be witnessed in Tab.\ref{physical-Twitter}. Compared with baselines, the superior performance of FingerSafe is consistent on real-world photoed images (\textbf{27.5\%} protection improvement over the best baseline). We also stress that FingerSafe achieves the best performance when preprocessors are used, which is closest to real-world fingerprint recognition scenario \cite{sankaran2015ijcb}.

\subsection{Discussion}
In this section, we provide further evaluations of FingerSafe, including results under JPEG compression, execution time and comparison with ad-hoc protection methods such as blurring and pixelization.
\subsubsection{\textbf{Robustness over JPEG compression}}
Since social media platforms often use JPEG compression when uploading pictures for space saving, we hereby evaluate the performance of FingerSafe under JPEG compression \cite{wallace1992jpeg} (also known as an adversarial defense method \cite{das2017jpeg}). Specifically, we first generate the protected images on a ResNet50 backbone, then compress the images using JPEG compression with different quality (lower quality means higher compression ratio). Finally, we feed the images into another ResNet-50 model with MHS as preprocessor. 
\begin{figure}
    \centering
    \includegraphics[scale=0.33]{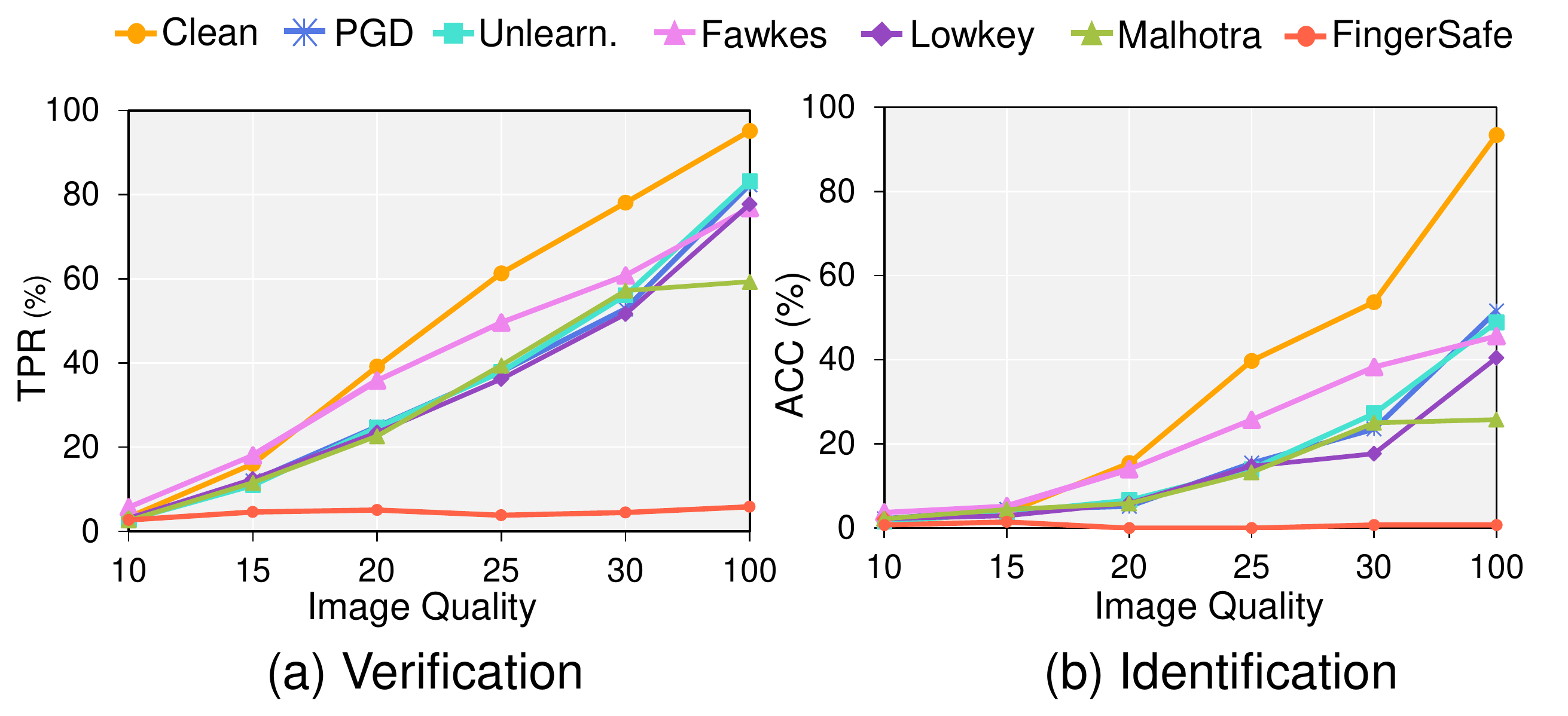}
    \caption{Results of FingerSafe and other baselines under different JPEG quality, lower quality means higher JPEG compression rate. Evaluated on both identification and verification task.}
    \label{compress}
\end{figure}
The results in Fig.\ref{compress} demonstrate that the protection of FingerSafe is robust across all image qualities in different JPEG compression rates (hackers succeed at most \textbf{1.47\%} probability). Moreover, from the trend of clean image and baselines, we conclude that the drop in accuracy in baseline methods is probably because of decreased image quality, instead of a stronger attacking capability.

\subsubsection{\textbf{Comparison of execution time}}
Protection in social media must be fast to ensure satisfactory user experience. We report the average execution time of protecting a single image using FingerSafe and baseline methods in Tab. \ref{tab-time}. According to Programs Wiki\footnote{https://programs.wiki/wiki/startup-optimization-of-android-performance-optimization.html}, the response speed of 2-5 seconds is acceptable for users. The execution time of FingerSafe satisfy this constraint.

\begin{table}[h]
\centering
\caption{Execution time of protecting a single image. The execution time of FingerSafe is acceptable on android applications.}\label{tab-time}
\scriptsize
\setlength\tabcolsep{2pt}
\begin{tabular}{@{}ccccccc@{}}
\toprule
Method              & PGD  & Unlearn.   & Fawkes & Lowkey & Malhotra & FingerSafe \\ \midrule
protection time (s) & 2.21 & 4.93       & 12.33  & 13.89  & 4.72    & 3.69       \\ \bottomrule
\end{tabular}
\end{table}


\subsubsection{\textbf{The effect of ad-hoc protection methods}}
We use experiments to support our claim that ad-hoc protection methods is ineffective. Specifically, we use two kind of ad-hoc protection methods, Gaussian blurring and pixelization. For Gaussian blurring, we smooth the fingerprint area using a Gaussian kernel with kernel size of 15, and adjust the strength of Gaussian blurring by changing the standard deviation $\sigma$ in Gaussian kernel. For pixelization, we pixelize different percentage of fingerprints to $10 \times 10$ mosaic. Then, we use an off-the-shelf deblurring algorithm, DeblurGAN \cite{kupyn2019deblurgan} and inpainting method, Deepfill \cite{shiri2019deepfill}, to reverse the blurring and pixelization process. The protection strength was selected via searching images with similar naturalness comparing with FingerSafe. For Gaussian protection, we select $\sigma$=\{1.5, 2, 2.5\} for protection. For pixelization, we select the pixelization percentage r=\{30\%, 40\%, 50\%\}. While pixelizing 30\% of the image yields low naturalness, we nonetheless select 30\% as the lowest blurring since further decreasing r leads to negligible protection. The results can be shown in Tab. \ref{ad-hoc-protection}. We make several conclusions as follows:


(1) FingerSafe outperform ad-hoc protection methods by a large margin in transferability (21.76\% improvement over the best performing method in blurring, 10.69\% improvement over the best performing method in pixelization). Moreover, while strong in protection, FingerSafe keeps natural appearance than all settings of pixelization and several strengths of blurring, demonstrating our viewpoint that the ad-hoc protection methods are either lack of transferability or visually unnatural. 

(2) Due to the ill-posed nature of the reconstruction problem, it is impossible for hacker to steal the fingerprint as easy as before.

(3) Gaussian noise do not transfer well to preprocessing-based methods since it do not perturb high-level semantics. Thus, the added noise are simply filtered out by preprocessors. In contrast, pixelization directly wipe out the details of fingerprint, thus achieving consistent transferability in all settings.

\begin{table}[]
\caption{The performance of two ad-hoc protection methods(\ie{blur and pixelization}) after simple de-noising methods(\ie{deblur and inpainting}). In this table, B($\sigma$) and P(r) represent different strength of Blurring and Pixelization, under standard deviation $\sigma$ and pixelization percentage r.}
\label{ad-hoc-protection}
\centering
\setlength\tabcolsep{3pt}
\scriptsize
\begin{tabular}{@{}ccccccccc@{}}
\toprule
            & \multicolumn{8}{c}{Veri. TPR(\%)}                                                                                               \\ \midrule
Method      & Clean & B(1.5) & B(2) & B(2.5) & P(30\%) & P(40\%) & P(50\%) & FingerSafe \\ \midrule
ResNet50    & 91.17 & 51.12       & 35.20     & 23.51       & 74.25                & 58.96                & 37.06                & 0.00       \\ \midrule
HG-ResNet50 & 88.06 & 70.77       & 67.04     & 62.83       & 68.53                & 61.94                & 51.19                & 4.60       \\ \midrule
HG-Scat-L2  & 75.75 & 36.97       & 33.33     & 27.11       & 29.85                & 24.88                & 16.04                & 5.35       \\ \midrule
Naturalness & 6.20   & 4.00         & 3.60       & 3.04           & 2.16                & 2.05                  & 1.78                 & 3.56        \\ \bottomrule
\end{tabular}
\end{table}

\subsection{Ablation studies}
In this section, we conduct ablation studies to verify the effect of different loss terms and hyperparameters, namely $\lambda$ that controls orientation field distortion loss $\mathcal{L}_{O}$ and $\gamma$ that controls local contrast suppression loss $\mathcal L_C$.

\subsubsection{\textbf{Effect of different loss terms}}
We conduct ablation studies to better understand the contributions of our two main loss terms, \ie, the orientation field distortion loss and local contrast suppression loss. We argue that orientation field distortion loss $\mathcal{L}_{O}$ mainly contributes to transfer attack in FingerSafe, while local contrast suppression loss $\mathcal L_C$ provides the natural appearance. To prove these views, we conduct an experiment by exploring different loss term combinations. On the basis of fixed $\mathcal{L}_{adv}$, we optimize FingerSafe with loss function $\mathcal{L}_{O}$, $\mathcal{L}_{C}$ and $\mathcal{L}_{O}+\mathcal{L}_{C}$ respectively. As shown in Tab.\ref{ablation}, the accuracy shows a significant drop under $\mathcal{L}_{O}$ setting (\ie, in MHS-ResNet, \textbf{0\%} under $\mathcal{L}_{O}$ and \textbf{0.74\%} under $\mathcal{L}_{O}+\mathcal{L}_{C}$ compared with \textbf{25.00\%} under $\mathcal{L}_{C}$). We evaluate naturalness using the same setting in section 4.3. The naturalness is restored by $\mathcal{L}_{C}$ (\ie, \textbf{2.73} under $\mathcal{L}_{O}$ and \textbf{4.75} under $\mathcal{L}_{O}+ \mathcal{L}_{C}$, p$\textless$.001, the improvement is \emph{significant}\footnote{We use nonparametric Wilcoxon signed-rank test to evaluate if the improvement of $\mathcal{L}_{C}$ is at significant level.}). Experiment results demonstrate our claim that $\mathcal{L}_{O}$ and $\mathcal{L}_{C}$ achieved their designed goal.

\begin{table}[t]
\caption{The ablation study of different loss terms when protecting fingerprint identification models. We set $\lambda$ and $\gamma$ as $10^{2}$ and $5 \times 10^{2}$ respectively. Naturalness is evaluated by the same method described in section 4.3. All components in FingerSafe achieved their desired goal.}
\centering
\setlength\tabcolsep{10pt}
\scriptsize
\begin{tabular}{@{}ccccc@{}}
\toprule
\multirow{2}{*}{Model} & \multicolumn{4}{c}{Iden. ACC (\%)}                                        \\ \cmidrule(l){2-5} 
                               &Raw & $\mathcal{L}_{O}$ & $\mathcal{L}_{C}$ & $\mathcal{L}_{O}+\mathcal{L}_{C}$ \\ \midrule
        ResNet                 & 94.12  & 0.00              & 0.74              & 0.00                              \\ \midrule
        MHS-ResNet             & 93.38  & 0.00              & 25.00             & 0.74                             \\ \midrule
        MHS-Scat-L2            & 83.82  & 2.21              & 54.41              & 3.68                              \\ \midrule
        Naturalness            & 5.56  & 2.73             & 5.35             & 4.75                             \\ \bottomrule
\end{tabular}
\label{ablation}
\end{table}

\begin{figure}[h]
\centering
\includegraphics[scale=0.27]{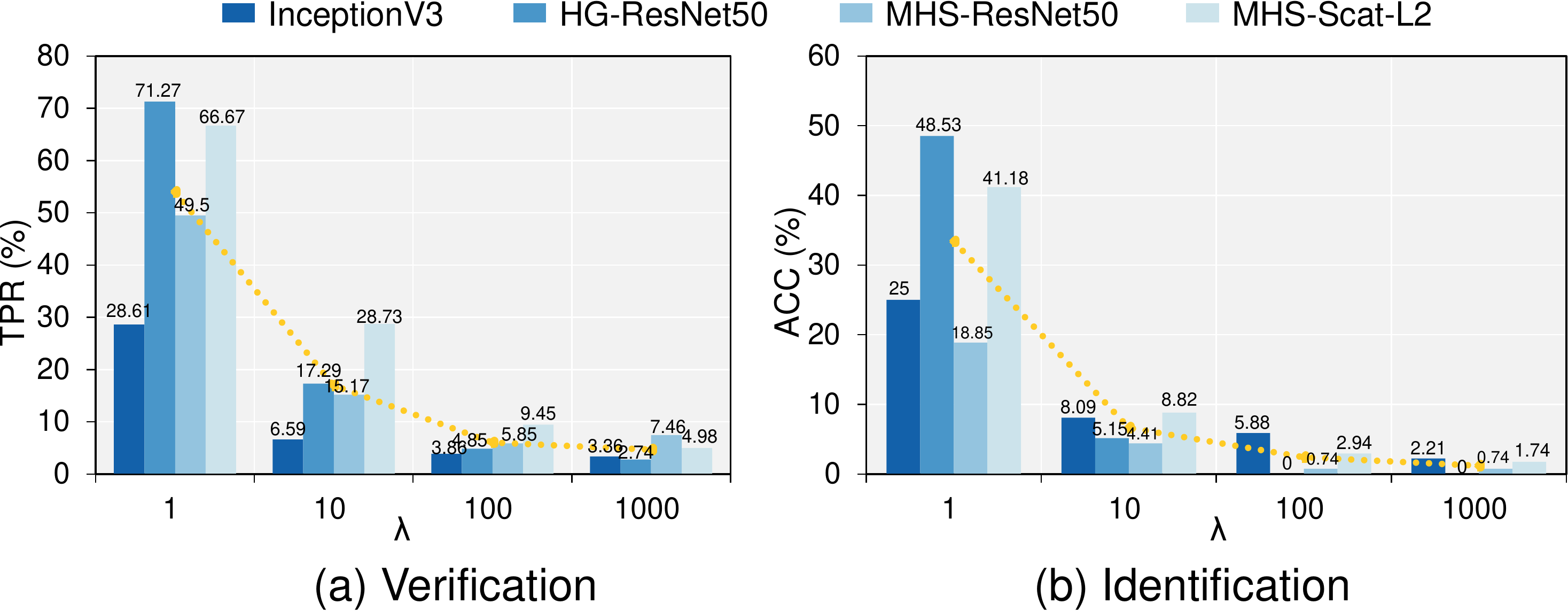}
\caption{Ablation on different $\lambda$ that controls $\mathcal{L}_O$. Dotted line indicates the average trend of accuracy. FingerSafe yields the best result with $\lambda \geq 100$. 
\label{ablation-lambda}}
\end{figure}

\subsubsection{\textbf{The effect of hyperparameter $\lambda$}}
Hyperparameter $\lambda$ controls the level of transferability. We evaluate the effectiveness of $\lambda$ on ResNet50 backbone and transfer it to different black-box models. We set the value of $\lambda$ to 1, 10, 100 and 1000, respectively. As illustrated in Fig.\ref{ablation-lambda}, the model accuracy first drops sharply with the increase of $\lambda$, then remains relatively stable as $\lambda$ further increases. The decline of model accuracy with preprocessing is more significant, from \textbf{71.27\%} to \textbf{4.85\%} as $\lambda$ increases from 1 to 100. This indicates that perturbing model-shared semantics is a more effective perturbation indicator for attacking black-box preprocessors. When $\lambda$ increase from 100 to 1000, accuracy increased slightly, meaning the perturbation induced by $\mathcal{L}_O$ have saturated.

\begin{figure}[h]
\centering
\includegraphics[scale=0.27]{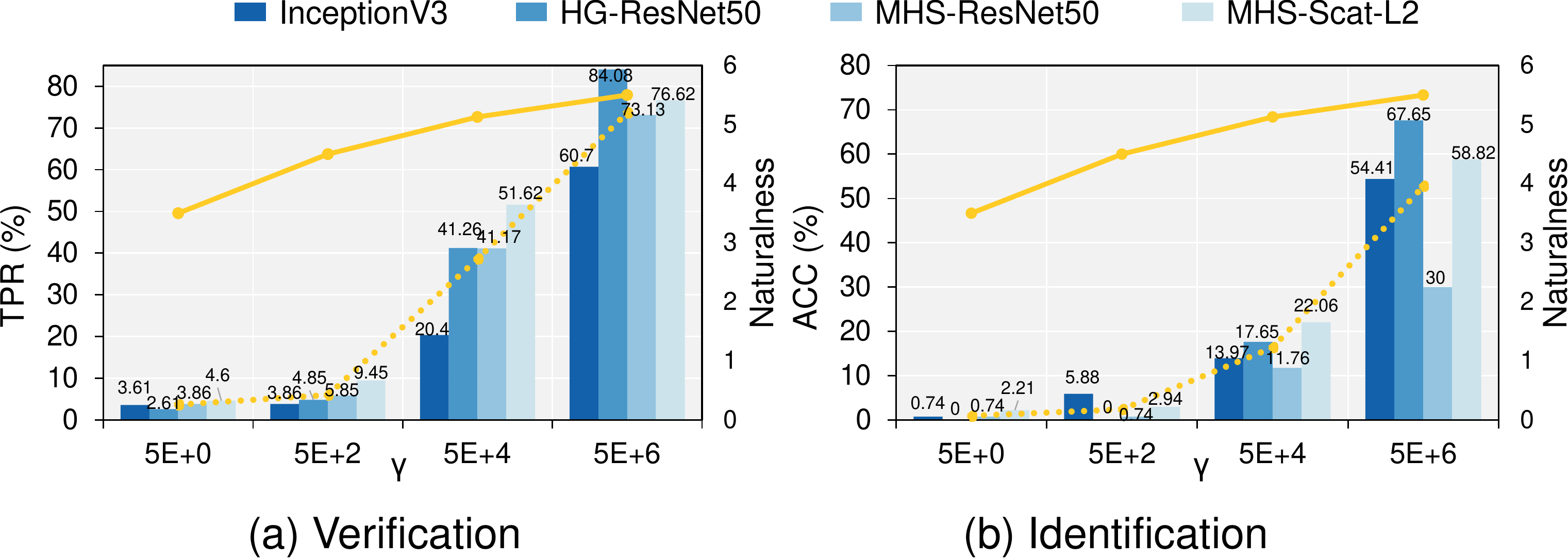}
\caption{Ablation on different $\gamma$ that controls $\mathcal{L}_C$ and controls the naturalness perceived by human. Dotted line indicates the average trend of accuracy, solid line shows the trend of naturalness as $\gamma$ increases.
\label{ablation-gamma}}
\end{figure}

\subsubsection{\textbf{The effect of different $\gamma$}}
Hyperparameter $\gamma$ controls the level of naturalness. We evaluate the effectiveness of $\gamma$ on ResNet50 backbone and transfer it to different black-box models. We set the value of $\gamma$ to 5, $5\times10^2$, $5\times10^4$ and $5\times10^6$, respectively. As illustrated in Fig.\ref{ablation-gamma}, the model accuracy first increase slightly with the increase of $\gamma$ (\eg, from 2.61\% TPR to 4.85\% TPR when $\gamma$ changes from 5 to $5\times10^2$), then increases largely as $\gamma$ further increases (\eg, from 4.85\% TPR to 41.26\% TPR when $\gamma$ changes from $5\times10^2$ to $5\times10^4$). When $\gamma = 5\times10^6$, all the protective patterns generated by orientation distortion loss $\mathcal{L}_O$ are simply wiped out. Naturalness improved stably with the increase of $\gamma$, but increase relatively faster at small $\gamma$. Based on following observations, we select $\gamma$ to be $5\times10^2$, retaining protection capability while maximally enhancing naturalness.



\section{Conclusion}


Biometric information (\eg, fingerprint) could be easily stolen from the social media images, causing irreversible hazards. To guard the fingerprint leakage, adversarial attack emerges as a new solution. However, existing works are either weak in black-box transferability or appear unnatural. To tackle the challenge, this paper proposes a hierarchical attack framework named FingerSafe by leveraging the low-to-high hierarchy of perception, which both attacks the high-level semantics and suppresses the low-level stimulus. Extensive experiments in both digital and real-world scenarios demonstrate FingerSafe outperforms other methods by large margins.

In the future, we will continue to improve our FingerSafe tool for fingerprint privacy protection, such that the recognition and processing speed is more suitable for real-world application. Moreover, we hope to collaborate with the industry to further conduct automatic fingerprint protection in social media applications, preferably by a fast and highly optimized engineering method. Finally, we wish to collaborate with social media industries to apply FingerSafe to protect fingerprint privacy leakage in real-world applications.





\bibliographystyle{IEEEtran}
\bibliography{ref}
%



%
\begin{IEEEbiography}[{\includegraphics[width=1in,height=1.25in,clip,keepaspectratio]{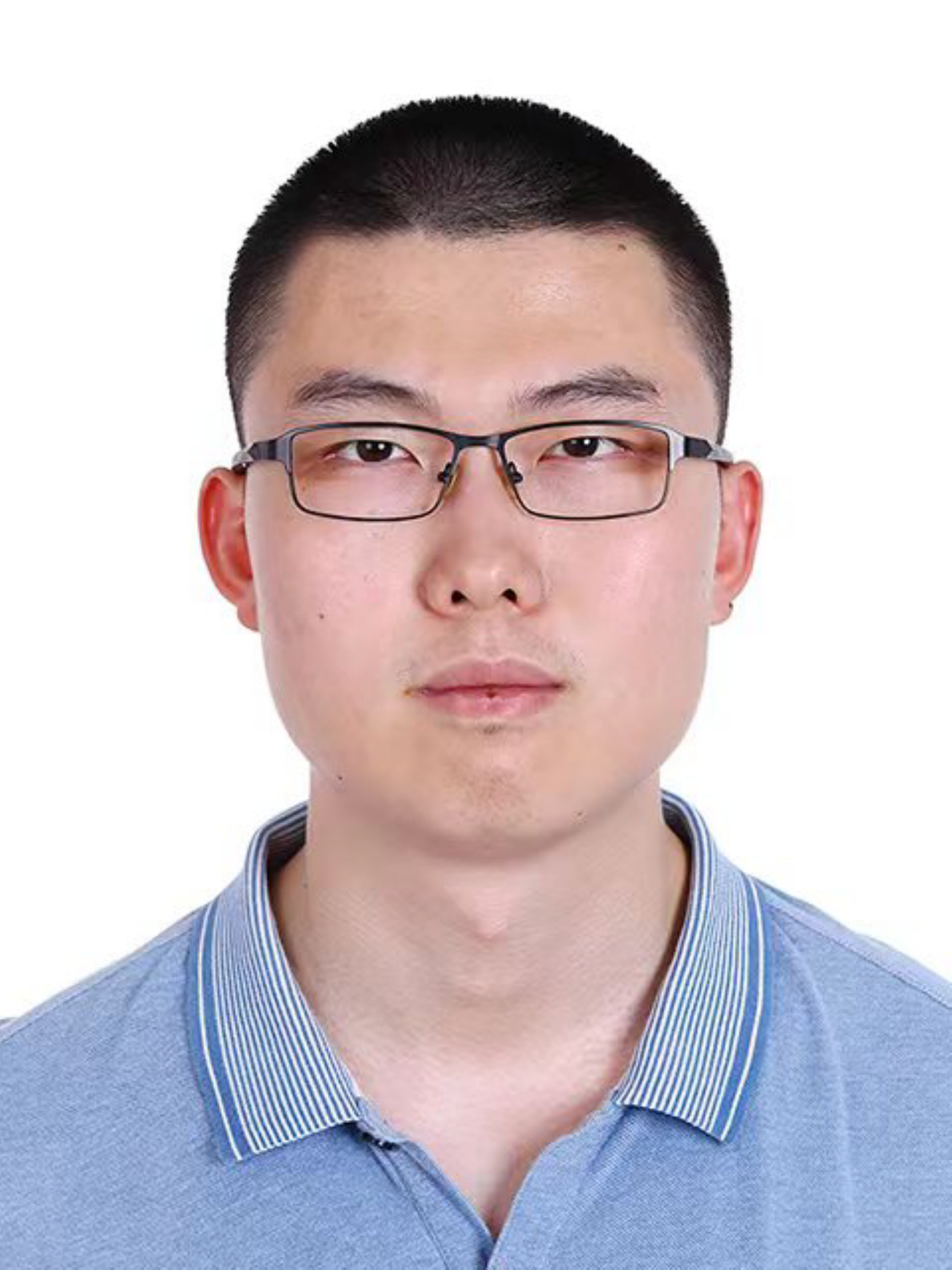}}]{Simin Li} (Graduate Student Member, IEEE)
received the B.S. in electronic engineering from Beihang University in 2016, where he is currently pursuing the Ph.D. degree with the School of Computer Science and Engineering. His current research interests include adversarial examples, trustworthy AI, AI for social good and computer vision.
\end{IEEEbiography}

\begin{IEEEbiography}[{\includegraphics[width=1in,height=1.25in,clip,keepaspectratio]{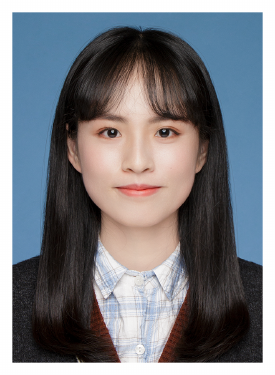}}]{Huangxinxin Xu}
received the BS degree from Wuhan University and is a postgraduate in Wuhan University. Her research interests include deep learning, computer vision and adversarial attacks.
\end{IEEEbiography}

\begin{IEEEbiography}[{\includegraphics[width=1in,height=1.25in,clip,keepaspectratio]{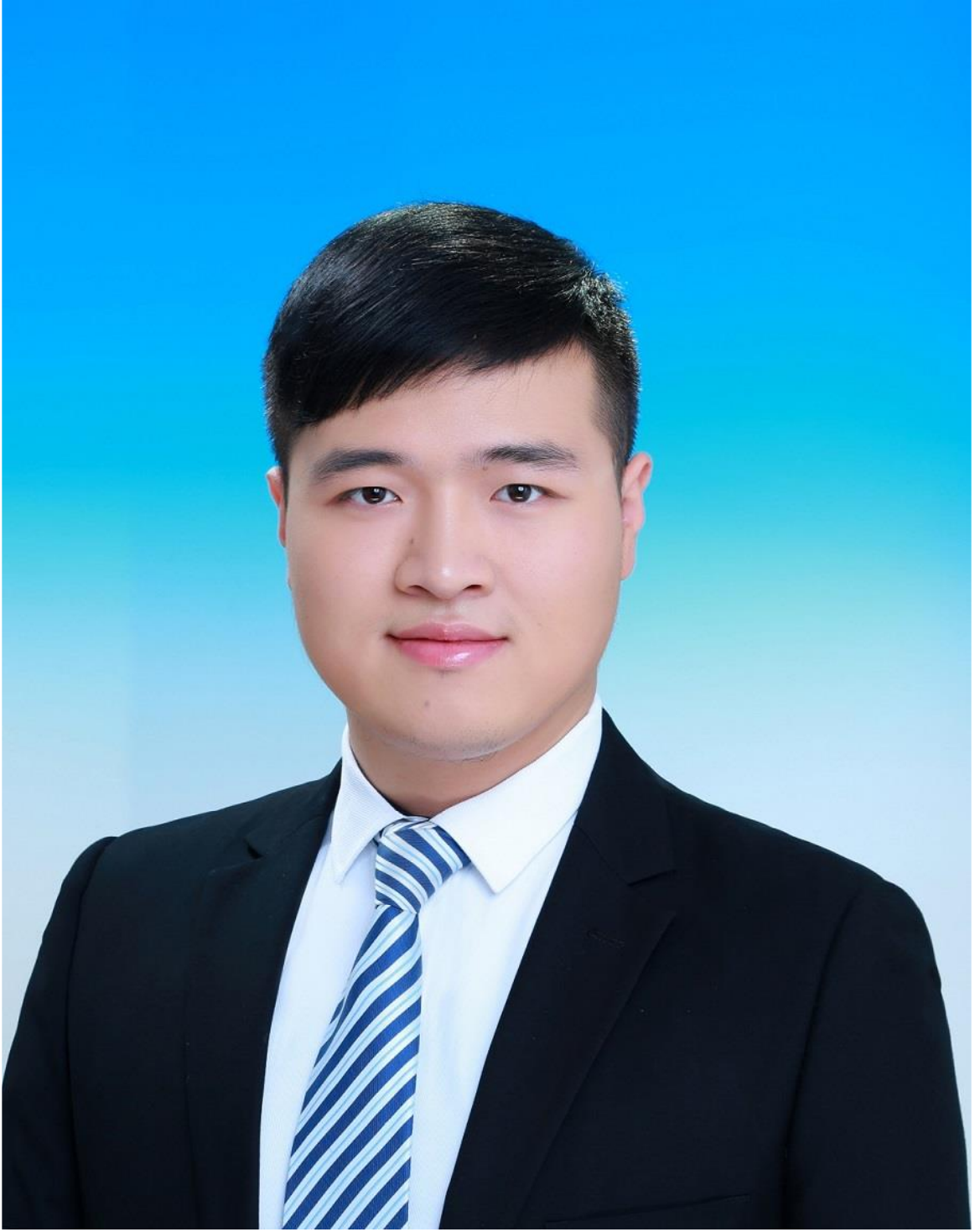}}]{Jiakai Wang} is now a Research Scientist in Zhongguancun Laboratory,a Beijing, China. He received the Ph.D. degree in 2022 from Beihang University, supervised by Prof. Wei Li and Prof. Xianglong Liu. Before that, he obtained his BSc degree in 2018 from Beihang University. His research interests includes Trustworthy AI in Computer Vision, which consists of the physical adversarial examples generation, adversarial defense and evaluation. 
\end{IEEEbiography}

\begin{IEEEbiography}[{\includegraphics[width=1in,height=1.25in,clip,keepaspectratio]{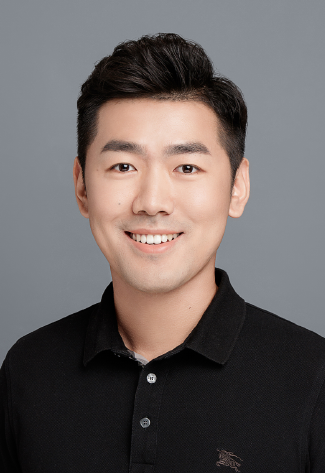}}] {Aishan Liu} (Member, IEEE) received the B.S., M.S., and Ph.D. degrees in computer science from Beihang University in 2013, 2016, and 2021, respectively. He is currently an Assistant Professor in Beihang University. His current research interests include adversarial examples and interpretable deep learning models, embodied agent, and computer vision. He has published over 30 research papers at top-tier conferences and journals.

\end{IEEEbiography}

\begin{IEEEbiography}[{\includegraphics[width=1in,height=1.25in,clip,keepaspectratio]{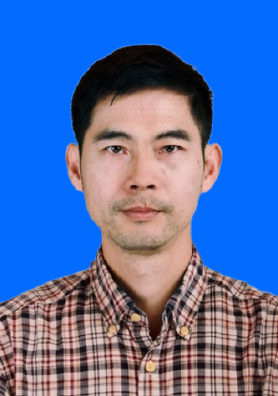}}]{Fazhi He} (Member, IEEE) received B.S. degree, M.S. degree and Ph.D. degree from Wuhan University of Technology. He was a Post-doctor Researcher in the State Key Laboratory of CAD\&CG at Zhejiang University, a Visiting Researcher in Korea Advanced Institute of Science \& Technology and a Visiting Faculty Member in the University of North Carolina at Chapel Hill. Now he is a Professor in School of Computer Science, Wuhan University. His research interest includes artificial intelligence, intelligent computing, computer graphics, computer-aided design, and co-design of software/hardware.
\end{IEEEbiography}

\begin{IEEEbiography}[{\includegraphics[width=1in,height=1.25in,clip,keepaspectratio]{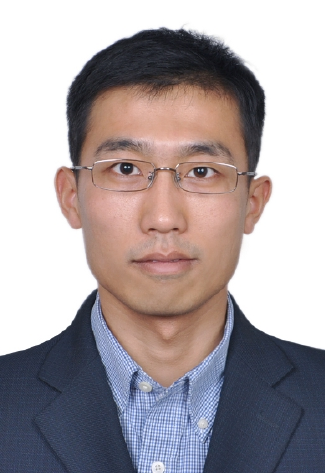}}]{Xianglong Liu}
(Member, IEEE) received the BS and Ph.D degrees in computer science from Beihang University, Beijing, China, in 2008 and 2014. From 2011 to 2012, he visited the Digital Video and Multimedia (DVMM) Lab, Columbia University as a joint Ph.D student. He is currently a full professor with the School of Computer Science and Engineering, Beihang University. He has published over 40 research papers at top venues like the IEEE TRANSACTIONS ON IMAGE PROCESSING, the IEEE TRANSACTIONS ON CYBERNETICS, the Conference on Computer Vision and Pattern Recognition, the International Conference on Computer Vision, and the Association for the Advancement of Artificial Intelligence. His research interests include machine learning, computer vision and multimedia information retrieval.
\end{IEEEbiography}

\begin{IEEEbiography}[{\includegraphics[width=1in,height=1.25in,clip,keepaspectratio]{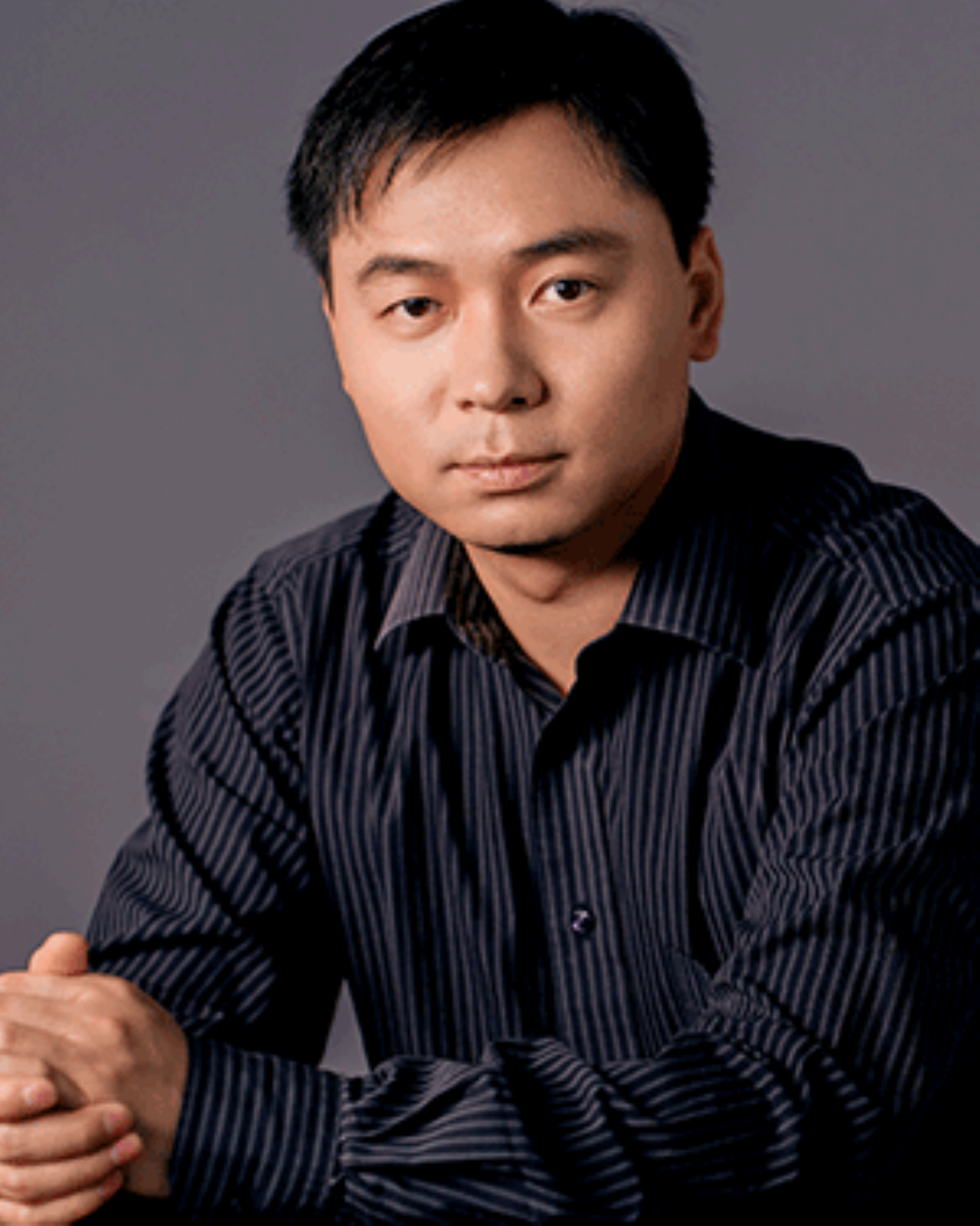}}]{Dacheng Tao}(Fellow, IEEE) is the Inaugural Director of the JD Explore Academy and a Senior Vice President of JD.com. He is also an Advisor and a Chief Scientist of the Digital Sciences Initiative, The University of Sydney. He mainly applies statistics and mathematics to artificial intelligence and data science. His research is detailed in one monograph and over 200 publications in prestigious journals and proceedings at leading conferences. He is a fellow of the Australian Academy of Science, AAAS, and ACM. He has received the 2015 Australian Scopus-Eureka Prize, the 2018 IEEE ICDM Research Contributions Award, and the 2021 IEEE Computer Society McCluskey Technical Achievement Award.
\end{IEEEbiography}








\end{document}